\documentclass[10pt,letterpaper]{article}
\usepackage[top=0.85in,left=2.75in,footskip=0.75in]{geometry}

\usepackage{amsmath,amssymb}

\usepackage{changepage}

\usepackage{textcomp,marvosym}

\usepackage{cite}

\usepackage{nameref,hyperref}

\usepackage[right]{lineno}

\usepackage[nopatch=eqnum]{microtype}
\DisableLigatures[f]{encoding = *, family = * }

\usepackage[table]{xcolor}

\usepackage{array}
\usepackage{makecell}
\usepackage{multirow}
\usepackage{adjustbox}
\usepackage{amsmath}
\usepackage{subcaption}
\usepackage{url}

\usepackage[percent]{overpic}

\newcolumntype{+}{!{\vrule width 2pt}}

\newlength\savedwidth

\raggedright
\usepackage[aboveskip=1pt,labelfont=bf,labelsep=period,justification=raggedright,singlelinecheck=off]{caption}

\makeatletter
\renewcommand{\@biblabel}[1]{\quad#1.}
\makeatother

\usepackage{lastpage,fancyhdr,graphicx}
\usepackage{epstopdf}
\fancyheadoffset[L]{2.25in}
\fancyfootoffset[L]{2.25in}
\begin{document}
\vspace*{0.2in}


\begin{flushleft}
{\Large
\textbf{An explainable hierarchical self attention-based approach for tremor detection in the time domain} 
}
\newline
\\
Timothy Odonga\textsuperscript{1},
Jeanne M. Powell\textsuperscript{1},
Mark Saad\textsuperscript{2},
Richa Tripathi\textsuperscript{2},
Christine D. Esper\textsuperscript{2},
Stewart A. Factor\textsuperscript{2},
Hyeokhyen Kwon\textsuperscript{1,3\ddag},
J. Lucas Mckay\textsuperscript{1,2,3\ddag}
\\
\bigskip
\textbf{1} Department of Biomedical Informatics, School of Medicine, Emory University, Atlanta, GA, USA
\\
\textbf{2} Jean and Paul Amos Parkinson’s Disease and Movement Disorders Program, Department of Neurology, School of Medicine, Emory University, Atlanta, GA, USA
\\
\textbf{3} Wallace H. Coulter Department of Biomedical Engineering, Georgia Institute of Technology, Atlanta, GA,  USA
\\
\bigskip

%
%

\ddag These authors also contributed equally to this work.

\textcurrency Current Address: Department of Biomedical Informatics, School of Medicine, Emory University, Atlanta, GA, USA 


\textpilcrow Membership list can be found in the Acknowledgments section.

* lucas@dbmi.emory.edu

\end{flushleft}


%
\section*{Abstract}
Tremor is a common movement disorder associated with conditions like Parkinson's disease and Essential tremor, traditionally diagnosed through expert clinician assessment.
Current automated detection methods rely on frequency-domain features informed by clinical expertise.
In this work, we present an explainable, two-stage hierarchical framework for tremor detection in the time domain that learns tremor patterns directly from 3D kinematic marker time-series data across entire tremor-provoking trials. 
Our framework combined a deep convolutional and long short-term memory network to learn tremor representations from short, discrete, non-overlapping time segments of kinematic time series data from trials, which are then processed by a vision transformer that models their long-term temporal dynamics of time segment features for trial (session) level classification.
Evaluated across nine body parts, the framework achieved F1-scores of $0.594-0.947$ depending on body parts (average: $0.765$), falling short of the frequency-domain state-of-the-art performance ($0.909$) while requiring minimal preprocessing. 
Attention weights and gradient-based class activation maps (Grad-CAM) identified time-domain features of tremor across body parts.
This proof of concept demonstrated the feasibility of data-driven time-domain modeling for tremor detection across anatomically diverse body parts, while reducing reliance on expert-engineered spectral features and providing posthoc interpretability of temporal and anatomical patterns of tremor.

\section*{Introduction}
\par Tremor is an involuntary, rhythmic, oscillatory movement of a body part that occurs across a range of neurological conditions, including Parkinson’s disease (PD) and essential tremor (ET)~\cite{bhatia2018consensus, deuschl1998consensus, testa2022}. 
Characterizing where, how, and under which conditions tremor appears is central to diagnosis and longitudinal assessment, yet in clinical practice, this characterization relies on expert observation by trained movement disorder neurologists, creating a bottleneck in access and scalability. 
Consequently, there is strong motivation to automate tremor assessment from movement data~\cite{pulliam2017continuous,heldman2014clinician,dorsey2018parkinson,maetzler2016clinical,de_machine_2023}. 

\par Clinical assessment of tremor is formalized in clinical rating scales such as the Movement Disorder Society Unified Parkinson's Disease Rating Scale Part III (MDS-UPDRS-III)~\cite{goetz_movement_2008} and The Essential Tremor Rating Scale (TETRAS) for ET~\cite{elble2016essential}. 
In these assessments, patients perform standardized tasks designed to elicit tremor under specific conditions, including rest (e.g., when quietly sitting, during walking, and without voluntary activation), posture (e.g., hands held outstretched in front of the body), and action (e.g., alternately pointing from the examiner’s finger to the patient’s nose, among other tasks)~\cite{deuschl1998consensus, bhatia2018consensus}. 
Tremor is evaluated separately across body regions and task conditions, and clinicians assign ordinal severity scores based on visual observation, which reflect the maximum observed amplitude within a brief examination period~\cite{bhatia2018consensus}. 
Tremor constancy is scored separately from amplitude, reflecting the clinical recognition that tremor may be continuous or intermittent within a single trial and that this distinction carries diagnostic weight~\cite{goetz_movement_2008}. 
The conditions under which tremor appears, its distribution across body segments, and its temporal behavior within a trial are all clinically meaningful dimensions of assessment~\cite{deuschl1998consensus, bhatia2018consensus}.

\par Automating this assessment requires reducing these multidimensional observations to computable features. 
Tremor frequency has historically served as the primary quantitative descriptor, alongside body distribution and provoking conditions~\cite{deuschl1998consensus,bhatia2018consensus}, because it provides a relatively stable and easily measurable summary of the oscillatory process~\cite{bhatia2018consensus}.
In PD, tremor typically occurs at a relatively narrow frequency band (roughly 4–6 Hz) and is thought to arise from a central oscillatory mechanism involving distributed basal ganglia–thalamo–cortical circuits~\cite{goetz_movement_2008}.
ET similarly exhibits characteristic frequency bands (typically 4–12 Hz), though at higher frequencies than PD and primarily during posture~\cite{deuschl1998consensus}.
Because this oscillation is relatively stable and narrow-band, frequency-based features (e.g., peak frequency, band power, and related spectral statistics) have been effective descriptors for identifying Parkinsonian tremor, and their clinical salience naturally motivated their adoption in automated detection pipelines \cite{samadi2020analysis,beuter1999,pang2020,luksys2018,brittain2015,meigal2012}.

\par Across sensing modalities including accelerometry~\cite{zeuner2003accelerometry}, 3D optical kinematics~\cite{marksaadDevelopmentTremorDetection2024} and pose recognition~\cite{friedrich2024video} among others~\cite{haubenberger2016transducer}, existing methods reflect this reliance on frequency representiation. 
Prior work has typically framed tremor detection as a classification problem, in which motion signals are analyzed over short windows, transformed into the frequency domain, and reduced to descriptive statistics characterizing oscillatory content. 
These engineered features are then aggregated and used as inputs to classifiers that predict tremor presence or absence.
This approach has proven effective for sustained, high-amplitude tremor, where spectral properties are relatively stable across a trial. 
However, it relies on an implicit stationary assumption~\cite{brown1997introduction, yates2025probability} that does not hold for intermittent tremor (i.e., tremor that emerges, re-emerges, or fluctuates during a trial). 

\par In clinical practice, expert neurologists are instructed to rate the worst tremor amplitude observed during a task, while constancy of tremor is scored separately on the MDS-UPDRS-III (e.g., rest tremor amplitude vs. constancy of rest tremor)~\cite{goetz_movement_2008}. 
However, automated methods that rely on aggregated spectral summaries average within-trial tremor variation over the entire recording, causing intermittent tremor to appear attenuated relative to the worst tremor amplitude. 
This can miss clinically meaningful fluctuations in tremor amplitude over time~\cite{de_machine_2023}. 


\par Furthermore, common steps in preprocessing pipelines can suppress physiologically meaningful temporal variation of tremor~\cite{de_machine_2023}.
For example, clipping fixed intervals from the beginning or end of recordings to exclude artifacts and instability associated with the start of trials may remove segments that may contain intermittent tremor~\cite{jeon2017high}.
These pipelines also typically involve several sequential steps: zero-phase low-pass filtering for noise reduction, Savitzky-Golay derivative filtering for marker velocity calculation, Butterworth band-pass filtering for tremor frequency isolation, normalization for scaling, and spectral decomposition via Welch's method or time-frequency decomposition via continuous wavelet transform for feature extraction~\cite{gallagher2020savitzky, marksaadDevelopmentTremorDetection2024,chen2025wavetremor}.
Such cascaded operations also introduce cumulative latency~\cite{oppenheim1999discrete}, limiting applicability to real-time or continuous monitoring scenarios.

\par Existing tremor detection methods span traditional signal processing and modern deep learning approaches, yet consistently rely on frequency-domain representations. 
Traditional clinical algorithms predominantly convert kinematic data to the frequency domain via Fast Fourier Transform prior to tremor classification~\cite{marksaadDevelopmentTremorDetection2024, haubenberger2016transducer, raethjen2004tremor, deuschl1995tremor, timmer2000cross}. 
Wavelet-based methods occupy an intermediate position, decomposing signals into spectro-temporal components through continuous wavelet transforms or Kalman filtering frameworks~\cite{jung2022imu, chen2025wavetremor, shahtalebi2019wake}, but most ultimately pass spectral image representations to classifiers~\cite{jung2022imu, chen2025wavetremor}.
Deep learning approaches, such as convolutional~\cite{kim2018wrist, sigcha2021automatic, san2020parkinson} and recurrent architectures~\cite{ribeiro2019bag, hssayeni2019, shahtalebi2020phtnet} have been applied directly to tremor time series data, but these methods still predominantly rely on frequency-domain transformations or aggregate statistics as inputs~\cite{kim2018wrist, san2020parkinson}. 
No prior work has applied a hierarchical deep learning framework in the time domain to learn tremor-relevant representations directly from kinematic time series data for tremor detection.

\par In this work, we propose a two-stage hierarchical framework to address this gap.
This work presents a proof-of-concept time-domain approach for tremor detection, offering a complementary method to existing frequency-domain pipelines.
In the first stage, a deep learning model~\cite{ordonez_deep_2016} learns tremor-relevant features from non-overlapping time segments of each trial directly from raw kinematic marker data, with minimal preprocessing. 
In the second stage, a classifier model aggregates the learned representations to produce a binary tremor-detection decision across the entire trial~\cite{suthaharan2016support, dosovitskiy2020image}. 
The framework also supports post-hoc interpretability, identifying which temporal windows and kinematic markers most influenced each detection~\cite{vaswani2017attention,selvaraju2017grad}.

\section*{Materials and methods}
\subsection*{Dataset}
\begin{table}[t!]
    \centering
    \caption{Nomenclature and description of the tremor-provoking tasks implemented in the Saad dataset~\cite{marksaadDevelopmentTremorDetection2024}}
    \begin{adjustbox}{max width=\linewidth}
    \begin{tabular}{ll}
        \hline
        \\
        Code & Task\\
        \\
        \hline
        sit-rest & Seated with arms placed on the thighs\\
        sit-arms-extended & \makecell[l]{Seated, with arms extended anteriorly and parallel to the floor}\\
        sit-UEopp & \makecell[l]{Seated, with arms in a ``T" pose parallel to the ground with\\fingers at each hand opposed}\\
        sit-point & \makecell[l]{Seated, performing a finger-to-nose pointing task with the\\indicated extremity}\\
        sit-spiral & \makecell[l]{Seated, performing a spiral movement with the indicated\\extremity (right or left)}\\
        std-rest & Standing with arms at sides\\
        std-arms-extended & Standing, with arms extended out parallel to the ground\\
        std-UEopp &\makecell[l]{Standing, with arms in a ``T" pose parallel to the ground\\with fingers of each hand opposed}\\
        walk-thru &\makecell[l]{Comfortable walking from one end to the other of the motion\\capture space}\\
        \hline 
    \end{tabular}
    \end{adjustbox}
    \label{tab:tremor_tasks}
\end{table}


\subsubsection*{Participants}

Data were derived from recordings of 52 patients evaluated for movement disorders in the Brain Health 3D Motion Capture Laboratory at Emory University. 
All data were collected under approval of the Internal Review Board (IRB) at Emory University (IRB \#00002688). 
Patients contributed one or more recording sessions corresponding to standardized tremor-provoking tasks.

\subsubsection*{Kinematic data acquisition}

Participants were prompted to perform a standardized set of tremor-provoking tasks while instrumented with 60 reflective markers placed at predefined anatomical locations. 
Each contiguous recording corresponding to a single tremor-provoking task is referred to as a session. 
A detailed list and description of the tremor-provoking tasks are provided in \autoref{tab:tremor_tasks}.
Kinematic data were captured using a 14-camera optical motion capture system (Motion Analysis Corporation, Rohnert Park, CA, USA) sampling at 120 Hz. 

\subsubsection*{Clinical labeling of tremor}

In parallel with kinematic acquisition, clinicians documented their observations of tremor in the patient’s clinical note (e.g., “mild, intermittent tremor in right hand”). 
These observations were converted into binary tremor labels (present vs absent) for each of 16 predefined body extremities (head, shoulders, thorax, pelvis, left/right hand, left/right foot, left/right proximal arm, left/right distal arm, left/right distal leg, and left/right proximal leg) and assigned at the session level. 
Only explicitly documented tremor present or absent observations were labeled; body parts not mentioned in the clinical note were treated as unlabeled to avoid assuming absence of tremor from lack of documentation.

\subsubsection*{Anatomical mapping for modeling}

Multiple markers were part of a single extremity as illustrated in~\autoref{fig:marker_placement}.
Detailed marker placement information for all body extremities is provided in ~\autoref{tab:trunk_markers},~\autoref{tab:arms_markers},and~\autoref{tab:legs_markers} reproduced from Saad \textit{et al.}\cite{marksaadDevelopmentTremorDetection2024}. 
These tables document the anatomical descriptions and coding conventions for kinematic markers in each extremity.

\subsubsection*{Label distribution}
Analyses were conducted on a subset of nine body extremities comprising both hands, both feet, the head, and four leg segments (left/right distal and proximal), totaling $1,750$ recordings from all $52$ subjects (77\% of the full labeled dataset).
The label distribution showed substantial class imbalance, with 714 tremor-present recordings (40.8\%) versus 1,036 tremor-absent recordings (59.2\%).
Hand tremor dominated the subset (280 and 279 tremor-present recordings for the left and right hands, respectively), while head tremor was moderately represented (52 recordings).
Leg and foot body extremities showed fewer tremor-present recordings, ranging from 8 to 30 recordings per extremity.
Other body extremities (e.g., shoulder, pelvis) were excluded from the analysis as the number of tremor present recordings ($\leq$ 5 recordings) was insufficient for supervised training.
The detailed distribution of tremor labels in the dataset is shown in~~\autoref{tab:label_distribution} and has been described previously in Saad \textit{et al.}~\cite{marksaadDevelopmentTremorDetection2024}. 



\subsection*{Clinical baseline algorithm}
\par At the Emory Brain Health 3D Motion Capture Laboratory, the current clinical standard for automated tremor detection relies on expert-engineered spectral features extracted from optical motion capture data, the gold standard for kinematic measurement.
This approach comprises two complementary algorithms: a velocity-based method and an amplitude-based method.
This approach is actively deployed for billable clinical assessments at the Emory Brain Health 3D Motion Capture Laboratory, where it has been iteratively refined over several years through real-world patient care workflows. 
Saad et al.~\cite{marksaadDevelopmentTremorDetection2024} evaluated this method on the same dataset.
For this study, we evaluated both algorithms and used the best-performing algorithm as the frequency-domain baseline.

\par The velocity-based algorithm applied zero-phase low-pass filtering (20 Hz) and Savitzky-Golay derivative filtering to displacement data to obtain smooth velocity estimates~\cite{gallagher2020savitzky}, computed power spectral densities using Welch's method~\cite{welch1967use}, and combined them via the Euclidean norm across the x, y, and z dimensions. 
Then, a log transformation and Savitzky-Golay smoothing were applied prior to peak detection~\cite{gallagher2020savitzky}.
Rule-based classification identified tremor via narrow, symmetric spectral peaks (bandwidth $<$2 Hz, center frequency $<$10 Hz), with features extracted independently per marker and aggregated using a winner-take-all approach that selected the marker with maximal tremor amplitude.

\par The amplitude-based algorithm applied 4th-order Butterworth high-pass filtering (2 Hz cutoff) to displacement data, computed single-sided frequency spectra via fast Fourier transform for each marker axis, and aggregated across all markers using a max operation to represent the most severe tremor at each frequency. 
Afterward, the aggregate spectrum underwent Savitzky-Golay smoothing before heuristic peak detection~\cite{gallagher2020savitzky}. 
Rule-based classification identified tremor via peaks within the neurologic frequency range (3.5-10 Hz) that exceed an empirically determined amplitude threshold (0.1 mm), with aggregation performed in the frequency domain prior to peak selection rather than via winner-take-all marker selection.
These spectral features were also evaluated with a Support Vector Machine (SVM) classifier using radial basis function kernels and 5-fold cross-validation~\cite{suthaharan2016support}.

\subsection*{Preprocessing}

Session data were read from \verb|.trc| files and transformed from world-centered to pelvis-anchored, body-centered coordinates using established kinematic mapping methods~\cite{cappozzo1995position, pham2017algorithm}.
Marker series in the global laboratory frame were translated to the pelvis center of mass, then rotated into the pelvis anatomical coordinate system using a rotation matrix constructed from anatomical markers (anterior superior iliac spine and sacrum), yielding right-anterior-vertical (R-A-V) coordinates that move with the pelvis segment.
The transformed trajectories were then partitioned into non-overlapping four-second windows~\cite{bulling_tutorial_2014}. 
Windows shorter than four seconds were zero-padded to maintain uniform input dimensions. 
The four-second window duration was selected based on prior work demonstrating that brief motor events can be reliably detected within time windows of this length ~\cite{bachlin_wearable_2010, kwon_explainable_2023}. 
Given that tremor labels were only provided at the session level in the Saad dataset, we applied the session-level label to each window within the session as a proxy label for supervised training.

\subsection*{Hierarchical window-to-session modeling framework}
\begin{figure*}[t!]
\centering
\includegraphics[width=0.9\linewidth]{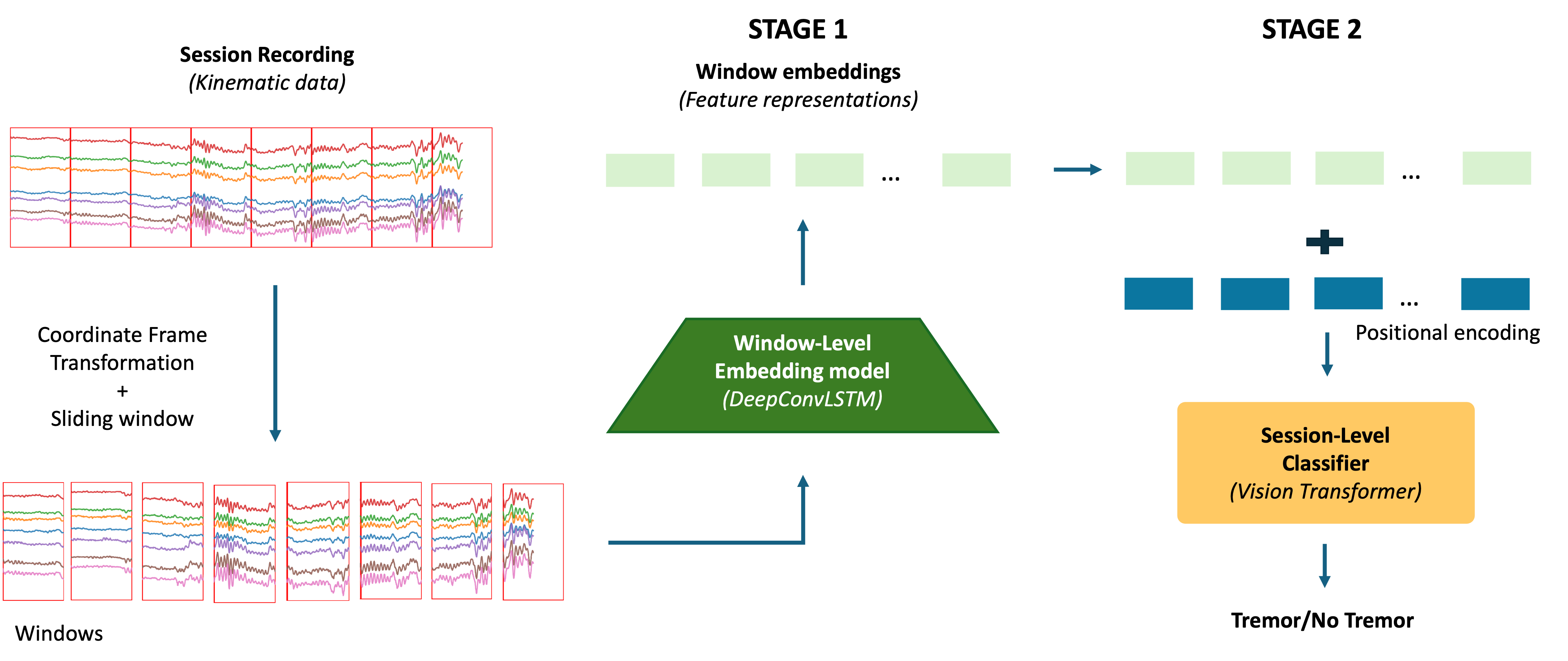}
\caption{Overview of the two-stage hierarchical framework for tremor detection. 
A recording session of multi-channel kinematic marker data in world-centered coordinates is converted to body-centered coordinates and segmented into sliding windows. 
Each window is processed by a DeepConvLSTM to generate window-level embeddings, which are combined with positional encodings and passed to a Vision Transformer (ViT) for session-level tremor classification by modeling temporal dependencies across the full sequence of windows.}
\label{fig:model_figure}
\end{figure*}

Our training procedure followed a hierarchical, two-stage modeling framework designed to separate local tremor representation learning from session-level decision making. 

\begin{enumerate}
    \item  A window-level model was trained to learn latent representations from fixed-length windows extracted from each recording session. 
    \item A session-level classifier was used to aggregate sequences of window-level embeddings produced in Stage~1 and generate a session-level tremor classification. 
\end{enumerate}

\autoref{fig:model_figure} provides a visual overview of the two-stage training framework and the data flow from raw kinematic windows to session-level classification.

\subsubsection*{Window-level embedding model}
The first stage learned fixed-dimensional representations of short-term tremor patterns from non-overlapping four-second windows. 
A deep convolutional and long short-term memory (DeepConvLSTM) architecture proposed by Ordonez and Roggen~\cite{ordonez_deep_2016} was used to capture both local temporal features and short-range dependencies within each window. 
The model processed multivariate marker trajectories through four one-dimensional convolutional layers (64 kernels each), followed by two LSTM layers (128 units each). 
Convolutional layers extracted localized temporal patterns, while LSTM layers modeled their temporal evolution across the window. 
A fully connected output layer produced window-level predicted labels during training. 
After training, embeddings were extracted from the hidden state at the final timestep of the second LSTM layer, yielding a fixed 128-dimensional representation for each window that served as input to the session-level model.

\subsubsection*{Session-level classifier}
The second stage aggregated window-level predicted labels to produce session-level tremor predicted labels. 
Three complementary approaches were evaluated, representing increasing levels of temporal modeling complexity.

First, a majority-vote approach assigned the session label based on the most frequent window-level predicted labels from the DeepConvLSTM model. 
This approach reflects a simple aggregation strategy analogous to summarizing the predominant tremor state observed during a task.

Second, an SVM classifier was trained using window-level embeddings aggregated by computing the mean values across each session. 
This method allowed the model to learn discriminative patterns across windows without explicitly modeling temporal order.

Third, a ViT encoder~\cite{dosovitskiy2020image} was used to model temporal relationships among window embeddings. 
In this framework, window embeddings were treated as sequential tokens with positional encodings to preserve their order within the session. 
A learnable class token integrated information across all windows to generate a session-level prediction. 
Conceptually, this approach parallels clinical assessment, in which brief tremor episodes are interpreted in the context of the entire task.

Positional encodings were added to the embeddings passed to the ViT to preserve their temporal order, and a learnable class token was prepended to aggregate information across all windows into a single representation for classification~\cite{dosovitskiy2020image, vaswani2017attention}. 
The token sequence was processed by a ViT encoder consisting of two transformer layers, each with multi-head self-attention (four heads, embedding dimension 128), followed by feed-forward networks. 
The final class token representation was passed to a fully connected classification head to generate the session-level tremor prediction.

\subsection*{Anatomical modeling configurations}
To enable body-part-specific tremor detection, we trained separate models for each anatomical region rather than a single unified model.
Each model was trained on data from a specific body part (e.g., hands, legs) across various tremor-provoking tasks. 
The input dimension of the window-level embedding model (DeepConvLSTM) was set to match the number of kinematic markers per extremity.
\autoref{tab:num_kinematic_markers} summarizes the number of kinematic markers in each body-part extremity.

\subsection*{Model performance and explainability}
\subsubsection*{Performance evaluation}

Model performance was evaluated using a five-fold user-independent nested cross-validation approach following the standard methods in machine learning (i.e., 80/20 train-test split ratio in each fold)~\cite{wainer2021nested, parvandeh2020consensus}. 
All data from each participant was assigned to a single fold, ensuring that no participant contributed data to both training and testing sets. 
This design enabled assessment of model generalization to previously unseen individuals. 
Within each training fold, data were further divided into training and validation sets, using an 80/20 split for model hyperparameter tuning. 

Macro F1-score was selected as the primary evaluation metric to account for the class imbalance between tremor and no-tremor labels across body parts~\cite{opitz2019macro}. 
Precision and recall were also reported to characterize false-positive and false-negative detection behavior relevant to clinical interpretation. 
To account for stochastic variability in model training, five iterations of cross-validation were conducted using different random seeds, and the mean performance, with 95\% confidence intervals, was reported. 


\subsubsection*{Intermittent tremor subgroup analysis}
To evaluate performance with respect to intermittent tremor, we conducted a targeted analysis of a subset of sessions from five participants who were clinically annotated as exhibiting intermittent tremor in the left hand, right hand, or left foot.
These five participants with 22 recordings were held out from training and used as a test set to evaluate model performance (using macro F1-score, precision, and recall) on intermittent tremor patterns.
This subset represented cases in which tremor occurred sporadically rather than continuously throughout a task (see panel (a) of ~\autoref{fig:intermittent_sustained}).

\subsubsection*{Model explainability}
\par Two complementary visualization approaches were applied to a high-performing iteration (complete 5-fold cross-validation run) to support the interpretation of model predictions.
First, attention weights from the transformer were extracted to identify temporal windows that most influenced session-level classifications~\cite{vaswani2017attention, bach2015, carion2020end}. 
Attention scores from the class token were averaged across transformer layers and heads to produce a single importance value per window~\cite{vaswani2017attention}. 
These scores were overlaid as color-coded window shading on time-series plots of per-marker displacement magnitudes, with red-bordered rectangles marking windows whose attention scores exceeded two standard deviations above the mean attention score (computed across all windows and sessions in the test fold), identifying windows most strongly prioritized by the model.

\par To quantify whether attention patterns differed between tremor present versus tremor absent recordings, we computed the variance of class-token attention scores across windows for each session.
To account for repeated measures arising from multiple sessions per subject, session-level variances were aggregated to the subject level by taking the median variance across each subject's correctly classified sessions within each class (tremor-present and tremor-absent separately). 
Only subjects with at least one correctly classified session in both the tremor-present and tremor-absent conditions were included, yielding a paired design. 
Sessions were pooled from folds with F1-score $\geq 0.80$ to ensure that attention weights reflected meaningful learned representations. 
The paired distributions of subject-level median variance were compared using a one-tailed Wilcoxon signed-rank test ($H_1$: tremor-present sessions exhibit greater window attention-score variance than tremor-absent sessions)~\cite{wilcoxon1992individual}.

\par Second, gradient-based class activation mapping (Grad-CAM) was applied to the final convolutional layer of the window-level model to identify marker channels contributing most strongly to tremor detection within individual windows~\cite{selvaraju2017grad}.
This adaptation produced marker-level saliency maps indicating which kinematic features drove the model's predictions.

\section*{Results}

\begin{table*}[t!]
    \centering
    \small
    \caption{
      Model performance for the tremor detection task for the body-part-specific model across different body parts.
      The window-level model used was DeepConvLSTM model~\cite{ordonez_deep_2016}.
      Performance is reported as mean $\pm$ 95\% CI across five separate data folds.
      Non-overlapping confidence intervals indicate that the difference between the values is statistically significant ($p\leq0.05$).
      $N$ denotes the number of recordings with the number of subjects in parentheses.
      \textbf{Bold} text indicates which session-level classifier scored the highest performance per body part. 
    }
    \begin{adjustbox}{max width=\linewidth}
    \begin{tabular}{cccccc}
        \hline
        \\
        Body Part & \makecell{N} & \makecell{Session-level\\Classifier} & F1-score & Precision & Recall\\
        \\
        \hline
         &  & Majority & $0.832 \pm 0.200$ & $0.831 \pm 0.198$ & $0.843 \pm 0.208$\\
         L\_Hand & \makecell{517\\(49)} & SVM & $0.816 \pm 0.079$ & $0.817 \pm 0.084$ & $0.833 \pm 0.073$\\
         &  & ViT & \textbf{0.844} $\pm$ \textbf{0.154} & \textbf{0.849} $\pm$ \textbf{0.198} & \textbf{0.845} $\pm$ \textbf{0.073}\\
        \hline
         &  & Majority & \textbf{0.915} $\pm$ \textbf{0.182} & \textbf{0.921} $\pm$ \textbf{0.174} & \textbf{0.921} $\pm$ \textbf{0.159}\\
         R\_Hand & \makecell{521\\(51)} & SVM & $0.887 \pm 0.188$ & $0.888 \pm 0.188$ & $0.894 \pm 0.172$\\
         &  & ViT & $0.892 \pm 0.180$ & $0.895 \pm 0.182$ & $0.889 \pm 0.176$\\
        \hline
         &  & Majority & $0.892 \pm 0.193$ & $0.894 \pm 0.190$ & $0.898 \pm 0.199$\\
        Head & \makecell{128\\(24)} & SVM & $0.898 \pm 0.230$ & $0.904 \pm 0.221$ & $0.896 \pm 0.233$\\
        &  & ViT & \textbf{0.947} $\pm$ \textbf{0.112} & \textbf{0.959} $\pm$ \textbf{0.107} & \textbf{0.946} $\pm$ \textbf{0.107}\\
        \hline
        & & Majority & $0.689 \pm 0.307$ & $0.741 \pm 0.279$ & $0.708 \pm 0.295$ \\
        L\_Foot & \makecell{106\\(17)} & SVM & $0.682 \pm 0.303$ & $0.725 \pm 0.292$ & $0.690 \pm 0.298$ \\
        &  & ViT & \textbf{0.718} $\pm$ \textbf{0.299} & \textbf{0.760} $\pm$ \textbf{0.289} & \textbf{0.716} $\pm$ \textbf{0.302} \\
        \hline
        & & Majority & $0.581 \pm 0.318$ & $0.585 \pm 0.338$ & $0.641 \pm 0.324$ \\
        R\_Foot & \makecell{96\\(17)} & SVM & $0.572 \pm 0.322$ & $0.559 \pm 0.336$ & $0.607 \pm 0.310$ \\
        &  & ViT & \textbf{0.594} $\pm$ \textbf{0.338} & \textbf{0.600} $\pm$ \textbf{0.357} & \textbf{0.628} $\pm$ \textbf{0.322} \\
        \hline
        & & Majority & \textbf{0.721} $\pm$ \textbf{0.247} & \textbf{0.725} $\pm$ \textbf{0.238} & \textbf{0.768} $\pm$ \textbf{0.243} \\
        L\_Dist\_Leg & \makecell{102\\(22)} & SVM & $0.679 \pm 0.254$ & $0.686 \pm 0.258$ & $0.721 \pm 0.247$ \\
        &  & ViT & $0.619 \pm 0.223$ & $0.620 \pm 0.236$ & $0.667 \pm 0.227$ \\
        \hline
        & & Majority & \textbf{0.726} $\pm$ \textbf{0.278} & \textbf{0.768} $\pm$ \textbf{0.294} & \textbf{0.728} $\pm$ \textbf{0.270} \\
        R\_Dist\_Leg & \makecell{89\\(22)} & SVM & $0.725 \pm 0.286$ & $0.767 \pm 0.296$ & $0.719 \pm 0.281$ \\
        &  & ViT & $0.725 \pm 0.280$ & $0.768 \pm 0.294$ & $0.726 \pm 0.271$ \\
        \hline
        & & Majority & \textbf{0.790} $\pm$ \textbf{0.301} & \textbf{0.857} $\pm$ \textbf{0.253} & \textbf{0.785} $\pm$ \textbf{0.318} \\
        L\_Prox\_Leg & \makecell{102\\(22)} & SVM & $0.771 \pm 0.337$ & $0.846 \pm 0.284$ & $0.759 \pm 0.362$ \\
        &  & ViT & $0.766 \pm 0.314$ & $0.824 \pm 0.281$ & $0.761 \pm 0.343$ \\
        \hline
        & & Majority & $0.778 \pm 0.379$ & $0.780 \pm 0.375$ & $0.775 \pm 0.383$ \\
        R\_Prox\_Leg & \makecell{89\\(22)} & SVM & $0.780 \pm 0.374$ & $0.786 \pm 0.366$ & $0.778 \pm 0.380$ \\
        &  & ViT & \textbf{0.780} $\pm$ \textbf{0.374} & \textbf{0.786} $\pm$ \textbf{0.366} & \textbf{0.778} $\pm$ \textbf{0.380} \\
        \hline
         & & Majority & $0.769 \pm 0.068$ & $0.789 \pm 0.067$ & $0.785 \pm 0.059$\\
        \makecell{Average\\(all body parts)} & -- & SVM & $0.757 \pm 0.069$ & $0.775 \pm 0.071$ & $0.766 \pm 0.063$\\
         & & ViT & $0.765 \pm 0.077$ & $0.785 \pm 0.077$ & $0.773 \pm 0.068$\\
         & & \textbf{Best} & $\mathbf{0.782 \pm 0.071}$ & $\mathbf{0.803 \pm 0.071}$ & $\mathbf{0.791 \pm 0.066}$\\
        \hline
    \end{tabular}    
    \end{adjustbox}
    \label{tab:all_results}
\end{table*}

\begin{table*}[t!]
    \centering
    \small
    \caption{
      Model performance for the tremor detection task for the body-part-specific model for patients annotated with intermittent tremor.
      The window-level model used is a DeepConvLSTM.
      Performance is reported as mean $\pm$ 95\% CI across five separate data folds.
      Non-overlapping confidence intervals indicate that the difference between the values is statistically significant ($p\leq0.05$).
      $N$ denotes the number of recordings with the number of subjects in parentheses.
      \textbf{Bold} text indicates which session-level classifier scored the highest performance per body part. 
    }
    \begin{adjustbox}{max width=\linewidth}
    \begin{tabular}{cccccc}
        \hline
        \\
        Body Part & \makecell{N\\(recordings,\\subjects)} & \makecell{Session-level\\Classifier} & F1-score & Precision & Recall\\
        \\
        \hline
         &  & Majority & \textbf{0.813} $\pm$ \textbf{0.115} & \textbf{0.803} $\pm$ \textbf{0.119} & $0.850 \pm 0.140$ \\
         L\_Hand & \makecell{8\\(3)} & SVM & $0.794 \pm 0.141$ & $0.787 \pm 0.150$ & $0.849 \pm 0.167$ \\
         &  & ViT & $0.786 \pm 0.108$ & $0.765 \pm 0.105$ & \textbf{0.851} $\pm$ \textbf{0.133} \\
        \hline
         &  & Majority & \textbf{0.685} $\pm$ \textbf{0.253} & \textbf{0.710} $\pm$ \textbf{0.239} & \textbf{0.772} $\pm$ \textbf{0.206} \\
         R\_Hand & \makecell{14\\(5)} & SVM & $0.651 \pm 0.251$ & $0.663 \pm 0.183$ & $0.755 \pm 0.223$ \\
         &  & ViT & $0.644 \pm 0.229$ & $0.669 \pm 0.181$ & $0.726 \pm 0.180$ \\
         \hline
         &  & Majority & \textbf{0.723} $\pm$ \textbf{0.141} & \textbf{0.702} $\pm$ \textbf{0.141} & \textbf{0.815} $\pm$ \textbf{0.032} \\
         Hands & \makecell{22\\(5)} & SVM & $0.698 \pm 0.140$ & $0.678 \pm 0.124$ & $0.805 \pm 0.040$\\
         &  & ViT & $0.691 \pm 0.149$ & $0.671 \pm 0.131$ & $0.788 \pm 0.081$\\
        \hline
    \end{tabular}    
    \end{adjustbox}
    \label{tab:all_results_intermittent}
\end{table*}

\begin{figure*}[!ht]
    \centering
    
\begin{overpic}[width=\linewidth]{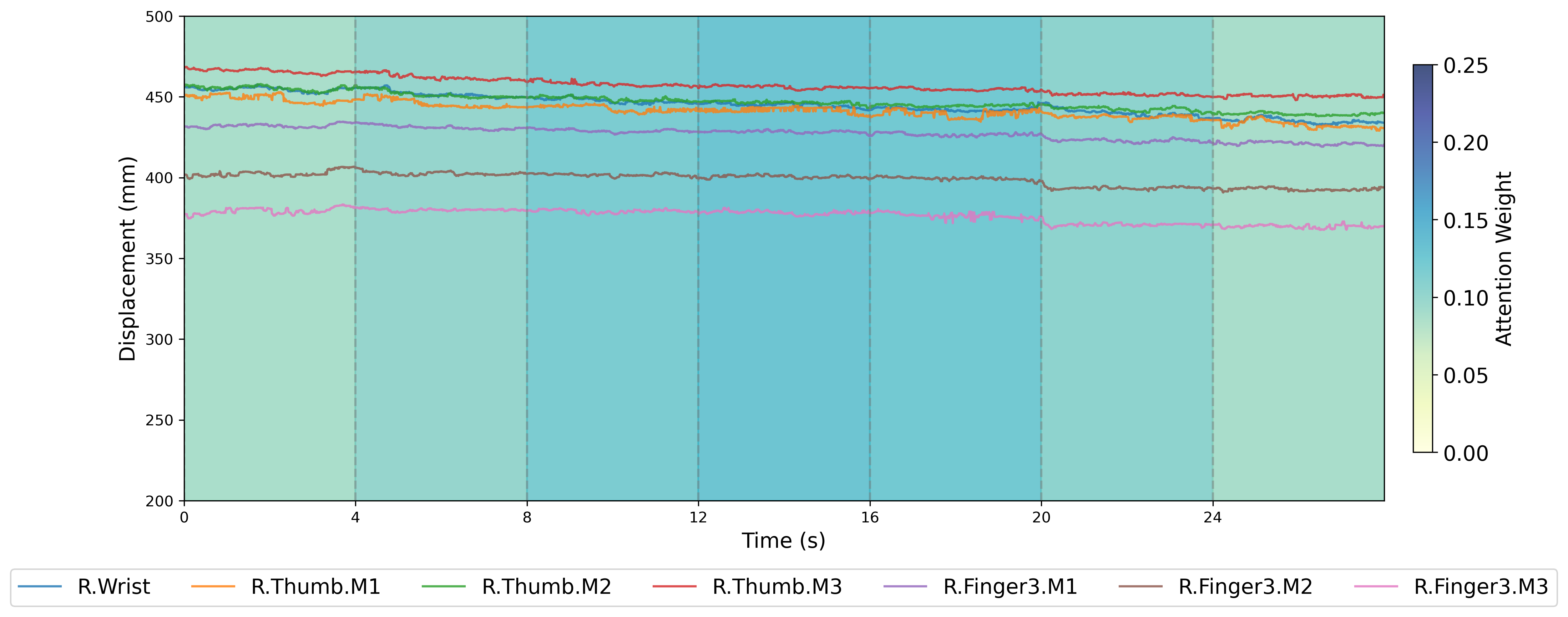}
    \put(2,40){\colorbox{white}{\large\textsf{\textbf{A}}}}
\end{overpic}

\vspace{2.5em}  

\begin{overpic}[width=\linewidth]{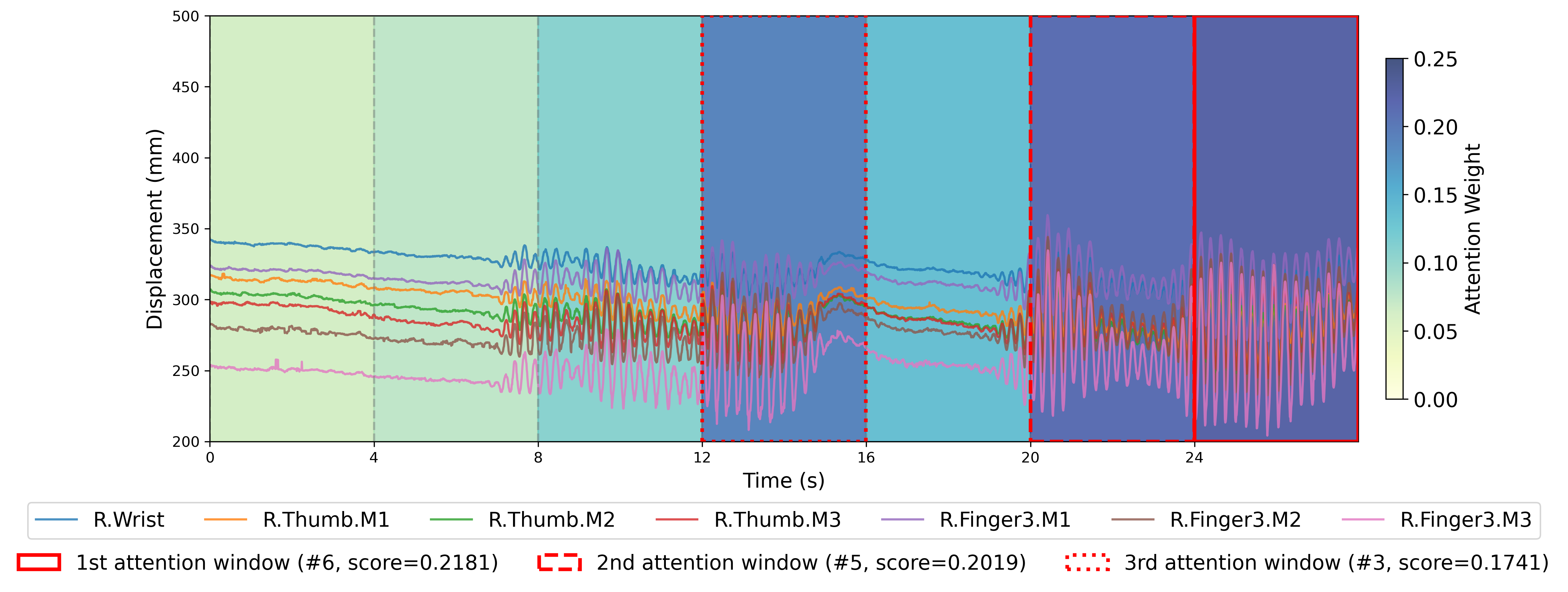}
    \put(2,38){\colorbox{white}{\large\textsf{\textbf{B}}}}
    \put(80,30){\huge\textsf{\textcolor{red}{$\star$}}}
\end{overpic}

\vspace{1.0em}  

    \caption{
    Visualization of attention weights and marker displacement time series for two correct model predictions during sit-spiral-left trials: (a) no tremor detected and (b) tremor detected. 
    For correct tremor prediction, the red borders indicate the top-3 windows the model attends to (windows 4, 6, and 7). Red star: The ViT classifier assigns the highest attention weights to the seventh window (24--28 s), capturing the largest tremor amplitude.
    For the correct no-tremor prediction, the model assigns relatively uniform weights (approximately 0.05) across all windows, correctly identifying the absence of visible tremor patterns.
    }
    \label{fig:attention_multiple_figures}
\end{figure*}
\begin{figure*}[!ht]
    \centering
    \begin{overpic}[width=\linewidth]{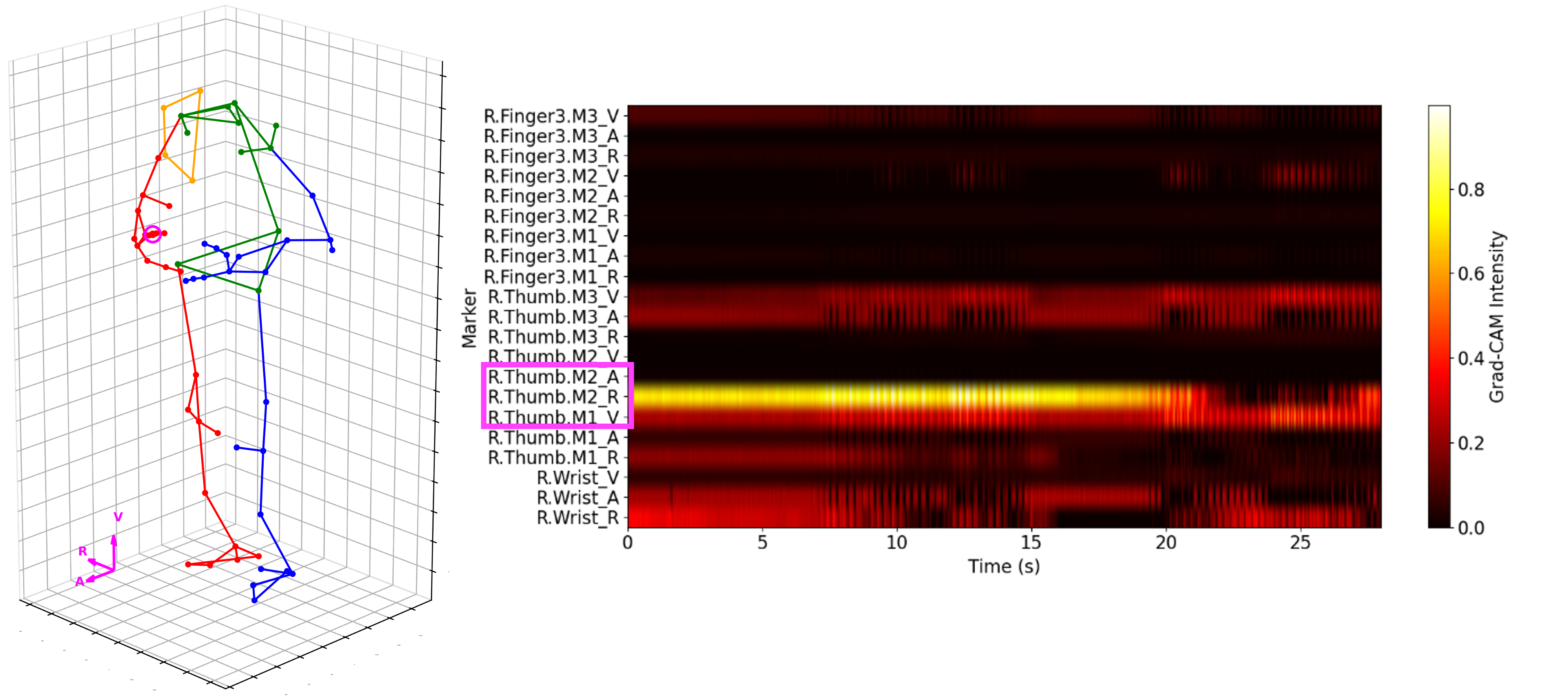}
    \put(-5,45){\colorbox{white}{\large\textsf{\textbf{A}}}}
    \put(35,45){\colorbox{white}{\large\textsf{\textbf{B}}}}
\end{overpic}

    \caption{
    Marker-level interpretability for a correct hand tremor prediction (panel (B) in ~\autoref{fig:attention_multiple_figures}). 
    (A) 3D kinematic skeleton during a standing T-pose trial with opposed-finger grip, with the dominant markers highlighted with a circle. 
    (B) Grad-CAM from the final convolutional layer of the window-level embedding model, localizing detection to the distal thumb markers (R.Thumb.M2, R.Thumb.M1), with emphasis on the right and vertical displacement components.
}

    \label{fig:gradcam_marker_example}
\end{figure*}

\begin{table*}[t!]
    \centering
    \small
    \caption{
      Average model performance for the different tremor detection approaches.
      Performance is reported as mean $\pm$ 95\% CI across five separate data folds from the best-performing iteration.
      Non-overlapping confidence intervals indicate that the difference between the values is statistically significant ($p\leq0.05$).
    }
    \begin{adjustbox}{max width=\linewidth}
    \begin{tabular}{lcccc}
        \hline
        \\
        Approach & Feature Type & F1-score & Precision & Recall\\
        \\
        \hline
         Saad et al~\cite{marksaadDevelopmentTremorDetection2024} & Frequency-domain & \textbf{0.909 $\pm$ 0.082} & \textbf{0.921 $\pm$ 0.081} & \textbf{0.923 $\pm$ 0.063} \\
         NeurDNet~\cite{shahtalebi2021deep} & STFT & 0.799 $\pm$ 0.054 & 0.834 $\pm$ 0.087 & 0.811 $\pm$ 0.043 \\
         Our approach & Time-domain & 0.782 $\pm$ 0.071 & 0.803 $\pm$ 0.071 & 0.791 $\pm$ 0.066 \\
         WaveTremor~\cite{chen2025wavetremor} & Wavelet & 0.758 $\pm$ 0.105 & 0.797 $\pm$ 0.091 & 0.763 $\pm$ 0.109\\
         Ensemble & \makecell{Time-domain,\\STFT, Wavelet} & 0.837 $\pm$ 0.056 & 0.845 $\pm$ 0.067 & 0.851 $\pm$ 0.047\\
         
        \hline
    \end{tabular}    
    \end{adjustbox}
    \label{tab:ablation_summary_table}
\end{table*}

\subsection*{Performance across body parts}
Performance varied across anatomical regions (F1-score: $0.594 - 0.947$), as shown in \autoref{tab:all_results}.
Head tremor detection achieved the highest performance, with the ViT session-level classifier reaching an F1-score of $0.947\pm0.112$. 
Upper extremity tremor detection was also robust, with right-hand models achieving F1-scores $0.887--0.915$ across all classifiers and left-hand models demonstrating moderate performance (F1: $0.816--0.844$).
Lower extremity detection showed greater variability.
Proximal leg segments achieved relatively strong performance (F1 = $0.790$), whereas distal foot tremor was most challenging, with both feet demonstrating the lowest performance (F1: $0.572--0.718$).
Classifier choice influenced performance differently across body parts. 
The ViT demonstrated the highest performance for head and left-hand tremor detection, whereas majority voting performed best for the right hand and several lower-extremity segments.
Overall, selecting the best session-level classifier per body part yielded an average F1-score of $0.782 \pm 0.071$ across all body parts.

\subsection*{Intermittent tremor subgroup analysis}
Performance on intermittent tremor cases was lower than overall detection performance, with F1-scores ranging from $0.644-0.813$ across body parts (~\autoref{tab:all_results_intermittent}). 
The majority voting classifier consistently achieved the highest performance in this subset, outperforming both SVM and ViT approaches.
Detection accuracy varied substantially by anatomical region. 
The model demonstrated the strongest performance on the left hand, while the right hand showed lower accuracy and greater variability.
When pooled across hands, the average F1-score was $0.704$ N=5 participants, 22 recordings) across all session-level classifiers, with majority voting achieving the best performance ($0.723\pm0.141$), followed by SVM ($0.698\pm0.140$) and ViT ($0.691\pm0.149$).

 

\subsection*{Model explainability}




\par The ViT's attention mechanism exhibited class-discriminative temporal focusing, concentrating on discrete windows of oscillatory activity in tremor-present sessions while distributing attention uniformly across windows in tremor-absent cases, as shown by the representative examples in ~\autoref{fig:attention_multiple_figures} of the sit spiral task performed on the left hand (Fold F1-score = $0.911$).
For the iteration from which the representative examples were drawn, correctly classified tremor-present recordings exhibited significantly greater attention score variance than tremor-absent recordings (W = 102, $p<0.001$, Wilcoxon signed rank test).
Among misclassified sessions, no significant difference in attention variance was observed, suggesting that classification errors coincide with a failure of the attention mechanism to preferentially weight tremor-relevant windows.



\par Marker-level Grad-CAM visualizations (Figure ~\ref{fig:gradcam_marker_example}) from the correctly classified tremor-present recording ((b) in ~\autoref{fig:attention_multiple_figures}) localized tremor features to anatomically plausible regions. 
The saliency maps revealed concentrated importance on the distal thumb markers (i.e., R.Thumb.M2\_R, R.Thumb.M1\_V, R.Thumb.M1\_A ), with periodic burst activations reflecting the rhythmic oscillatory nature of tremor during the standing T-pose task. 
The model emphasized the right and vertical displacement components, consistent with maintaining the opposed-finger grip posture with arms extended parallel to the ground.
In contrast, wrist markers (R.Wrist\_V, R.Wrist\_A, R.Wrist\_R) and finger markers (R.Finger3.M3\_V, R.Finger3.M2\_V) showed minimal contribution, indicating that the model localized tremor features to the distal thumb markers, exhibiting the most pronounced oscillatory motion during the posture.

\subsection*{Feature representation comparison}
To evaluate our choice of time-domain features, we compared them with frequency-based and time-frequency alternatives. 
We reimplemented the expert-derived frequency features from Saad \textit{et al.}~\cite{marksaadDevelopmentTremorDetection2024}, refined through years of clinical use, and two time-frequency approaches: the wavelet-based method by Chen \textit{et al.}~\cite{chen2025wavetremor} and the short-time Fourier transform (STFT) spectrogram-based CNN framework by Shahtalebi \textit{et al.}~\cite{shahtalebi2021deep}

~\autoref{tab:ablation_summary_table} shows performance across the five approaches. 
The frequency-based method achieved the highest F1-score ($0.909 \pm 0.082$), as expected, given that these features have been validated and refined through years of clinical practice. Our time-domain approach achieved $0.782 \pm 0.071$ without requiring domain-specific feature engineering. 
The two time-frequency approaches achieved $0.799 \pm 0.054$ for STFT spectrograms and $0.758 \pm 0.105$ for wavelets, with overlapping confidence intervals indicating no significant differences among the three learned approaches.
Combining the three feature types (time-domain, STFT, and Wavelet) in an ensemble improved performance to $0.837 \pm 0.056$, narrowing the gap with the frequency-based method (clinical baseline), with overlapping confidence intervals indicating no statistically significant difference.
Despite performing below the clinical baseline, all learned approaches had confidence intervals that overlapped with those of the frequency-based method.



\section*{Discussion}
\subsection*{Primary findings}
\par This work presented a data-driven hierarchical framework for tremor detection in the time domain that learned directly from raw kinematic marker data with minimal preprocessing (i.e., global-to-egocentric coordinate transformation and sliding window segmentation) and without relying on expert-driven spectral features. 
The two-stage approach, combining DeepConvLSTM for window-level feature extraction with ViT for session-level classification, achieved an average F1 Score of $0.765$ across nine body parts and tremor-provoking tasks. 
Attention-weight visualizations provided interpretable explanations of which temporal segments influenced tremor predicted labels, addressing transparency concerns in clinical machine learning applications. 
These findings provided proof of concept for data-driven deep learning approaches to tremor detection, demonstrating their feasibility as an alternative to methods relying on expert driven features and preprocessing pipelines~\cite{de_machine_2023}.

\par Our hierarchical deep learning framework achieved strong tremor detection performance for clinically critical body parts, with head tremor reaching F1 = $0.947$ and hand tremor achieving F1 = $0.844-0.915$. 
Performance varied substantially across the nine body parts evaluated, including bilateral hands, feet, head, and leg segments, with the lower extremities proving more challenging (F1 = $0.594$-$0.790$) due to subtle tremor manifestations and complex biomechanical patterns during different tremor-provoking tasks.
This difference in performance between the upper and lower extremities reflects the clinical reality that tremor occurrence and manifestation vary across body parts, with upper-body tremor providing more robust signals for detection~\cite{mahendran2022differentiation}. 
Notably, the optimal session-level classifier architecture varied by body part: ViT achieved superior performance on the head and feet, while simple majority voting outperformed more complex classifiers across several leg segments. 
This variation likely reflects substantial class imbalance across body parts, ranging from balanced cases in the hands to highly imbalanced cases in the distal and proximal legs (~\autoref{tab:label_distribution}), and suggests that aggregation strategies should be matched to the temporal characteristics of tremor in specific anatomical regions.

\subsection*{Comparison with prior work}

\par Our time-domain approach achieved an average F1-score of $0.765$ across nine body parts, significantly lower than $0.909$ for the frequency-domain methods implemented in Saad \textit{et al.}~\cite{marksaadDevelopmentTremorDetection2024} (p$<0.01$, Wilcoxon signed rank test), but comparable to time-frequency approaches: $0.799$ for the STFT spectrogram-based method\cite{shahtalebi2021deep}. and $0.758$ for the wavelet-based approach~\cite{chen2025wavetremor}, with no significant differences among the three learned methods.
While overall performance was lower than that of expert-engineered frequency features, our framework offers important trade-offs: it requires substantially less preprocessing and clinical-domain input, and learns tremor representations directly from raw kinematic signals rather than from engineered frequency features. 
The performance gap reflected the efficiency of expert-driven feature engineering, which directly encodes clinical knowledge about tremor characteristics, whereas data-driven approaches require larger datasets and more extensive training to learn comparable representations from raw signals~\cite{de_machine_2023, esteva2019guide, rajkomar2018scalable}.
Our explainability analysis also provided complementary insights: while frequency-domain SHAP (SHapley Additive exPlanations) identifies the frequency at which tremor occurs~\cite{marksaadDevelopmentTremorDetection2024, lundberg2017}, our attention mechanisms and Grad-CAM reveal when tremor emerges (temporal windows) and which kinematic markers contribute to its detection~\cite{vaswani2017attention, selvaraju2017grad}.
The comparable performance between our time-domain approach and the wavelet-based approach suggests that wavelet decomposition offered a limited advantage over direct temporal modeling for tremor detection across body parts, though both learned methods fell short of expert-engineered features.

\subsection*{Intermittent tremor subgroup analysis}

\par Intermittent tremor proved substantially more challenging to detect, with F1-scores decreasing by 0.1-0.2 points across body parts. 
This performance gap likely reflects both the inherent difficulty of detecting intermittent tremor episodes and the limitations of session-level proxy labeling, which may have introduced label noise, as many windows may have inherited tremor labels even when they contain no actual tremor activity. 
Notably, simple majority voting consistently outperformed ViT in these cases, suggesting that learning temporal dependencies via attention may be sensitive to label noise, whereas aggregating independent window predicted labels is more robust. 
Furthermore, transformer-type models are data-hungry and require large amounts of training data to achieve strong performance~\cite{liu2021efficient, wen2022transformers}, and the limited number of intermittent tremor sessions (N=5 participants, 22 recordings) may have constrained ViT's ability to learn meaningful temporal dependencies in this subgroup. 
Fine-grained temporal annotations, such as window-level tremor labels or onset and offset timestamps within sessions, would address the labeling limitation and potentially improve the detection of these clinically relevant edge cases.


\subsection*{Clinical implications}
The time-domain approach achieved performance within ~$10$\% of the frequency-domain clinical baseline algorithm despite data imbalance across extremities, suggesting potential as a complementary method to standard frequency-domain approaches in clinical workflows. 
Modeling temporal patterns with explainability at both the marker level (via Grad-CAM) and the temporal level (via attention weights) may offer additional clinical utility in differentiating tremor types, such as PD from ET~\cite{thanawattano2015temporal}.
This temporal and spatial resolution could supplement current clinical workflows, enabling structured documentation of when tremor occurred during a session and which body parts were most affected, thereby complementing clinicians' notes.

\subsection*{Limitations and future work}
\par Our evaluation was restricted to a single dataset from one clinical center~\cite{marksaadDevelopmentTremorDetection2024}, limiting the generalizability of our findings across patient populations, motion capture systems, and clinical settings. 
Future work should validate this approach on multi-site datasets with diverse patient populations and acquisition protocols to establish the robustness of time-domain deep learning for tremor detection.

\par The presence of substantial label imbalance with some body parts severely constrained performance for anatomical locations where tremor rarely manifests. 
The right foot achieved the lowest F1-score of $0.594$ with substantially fewer tremor-positive training sessions compared to the head and hands. 
This reflected the clinical reality that tremor prevalence varies across body parts, but it also revealed a fundamental challenge in applying supervised learning to such problems. 
Future work may explore data augmentation~\cite{de2024impact}, semi-supervised learning~\cite{huynh2022semi}, or transfer learning from high-prevalence to low-prevalence body parts to improve model performance in underrepresented anatomical regions~\cite{gupta2020transfer}.

\par Our evaluation on intermittent tremor detection was constrained by only five subjects with hand recordings, making it difficult to draw robust conclusions from this subgroup. 
Future work will prioritize collecting larger cohorts of patients with intermittent tremor to enable more robust evaluation of the framework's performance on this clinically relevant subpopulation.

\par Coarse session-level labels provide no information on when tremor occurred in a session. 
Our proxy-labeling of tremor during window-level training assigned the same label to all windows within a session, even though tremor may occur in only a subset of those windows. 
Fine-grained temporal annotations that mark the exact onset and offset of tremor would provide more accurate supervision and potentially improve the detection of low-amplitude, intermittent episodes.

\par Formulating tremor detection as a binary classification problem (i.e., present versus absent) in a session discards rich clinical information that could enhance diagnostic utility. 
Clinicians assess severity, frequency, characteristics, amplitude fluctuations, and contextual factors, while our framework reduces this multidimensional evaluation to a single binary decision~\cite{goetz_movement_2008, bhatia2018consensus}. 
Future work will extend the framework to multi-class severity classification using conformal prediction, thereby providing clinicians with both severity-class predictions and statistically valid uncertainty bounds~\cite{angelopoulos2023conformal}.

\section*{Conclusion}
Tremor is a common movement disorder associated with conditions like PD and ET, traditionally diagnosed through expert clinician assessment.
Current automated tremor detection methods predominantly rely on frequency-domain or time-frequency representations, leaving a methodological gap for pure time-domain temporal modeling approaches that could better capture tremor variability.
This work presents a proof-of-concept hierarchical framework for time-domain tremor detection that learns directly from raw kinematic data without requiring expert feature engineering.
Our approach achieved F1-scores of $0.765$ across nine body parts, comparable to time-frequency methods ($0.799$ for STFT spectrograms, $0.758$ for wavelets) but lower than frequency-domain methods ($0.909$) refined through decades of clinical use.
By modeling temporal dynamics through window-level feature extraction and session-level sequence modeling, our approach captured tremor patterns across nine body parts and provided dual explainability through attention weights (identifying important temporal windows) and GradCAM (identifying critical kinematic markers).
Our results demonstrate that temporal modeling could offer a complementary alternative to frequency-domain approaches by preserving information about when and where tremor occurs. 
Given the limitations of the lack of window-level labels and data imbalance across body parts, these results suggest that temporal approaches with window-level annotations and sufficient data could match frequency-domain performance.
Future work incorporating fine-grained temporal annotations could bridge the remaining performance gap relative to expert-engineered features while preserving the benefits of data-driven temporal representation learning, thereby advancing automated tremor assessment.

\bibliography{ref}

@article{deuschl1998consensus,
  title = {Consensus statement of the Movement Disorder Society on Tremor. Ad Hoc Scientific Committee},
  author = {Deuschl, G{\"u}nther and Bain, P. and Brin, M.},
  journal = {Movement Disorders},
  year = {1998},
  volume = {13},
  number = {Suppl 3},
  pages = {2--23},
  doi = {10.1002/mds.870131303},
  pmid = {9827589}
}

@incollection{testa2022,
    author = {Testa, Claudia M. and Haubenberger, Dietrich and Patel, Margi and Caughman, Christopher Y. and Factor, Stewart A.},
    isbn = {9780197529652},
    title = {Tremor in Medicine and Other Secondary Tremors},
    booktitle = {Tremors},
    publisher = {Oxford University Press},
    year = {2022},
    month = {09},
    doi = {10.1093/med/9780197529652.003.0007},
    url = {https://doi.org/10.1093/med/9780197529652.003.0007},
    pages = {75--84},
}

@article{zeuner2003accelerometry,
  title   = {Accelerometry to distinguish psychogenic from essential or parkinsonian tremor},
  author  = {Zeuner, K. E. and Shoge, R. O. and Goldstein, S. R. and Dambrosia, J. M. and Hallett, M.},
  journal = {Neurology},
  year    = {2003},
  volume  = {61},
  number  = {4},
  pages   = {548--550},
  doi     = {10.1212/01.wnl.0000076183.34915.cd},
  pmid    = {12939436}
}

@article{bhatia2018consensus,
  title={Consensus Statement on the classification of tremors. from the task force on tremor of the International Parkinson and Movement Disorder Society},
  author={Bhatia, Kailash P and Bain, Peter and Bajaj, Nin and Elble, Rodger J and Hallett, Mark and Louis, Elan D and Raethjen, Jan and Stamelou, Maria and Testa, Claudia M and Deuschl, Guenther and others},
  journal={Movement disorders},
  volume={33},
  number={1},
  pages={75--87},
  year={2018},
  publisher={Wiley Online Library}
}

@article{goetz_movement_2008,
  title={Movement Disorder Society-sponsored revision of the Unified Parkinson's Disease Rating Scale (MDS-UPDRS): scale presentation and clinimetric testing results},
  author={Goetz, Christopher G and Tilley, Barbara C and Shaftman, Stephanie R and Stebbins, Glenn T and Fahn, Stanley and Martinez-Martin, Pablo and Poewe, Werner and Sampaio, Cristina and Stern, Matthew B and Dodel, Richard and others},
  journal={Movement disorders: official journal of the Movement Disorder Society},
  volume={23},
  number={15},
  pages={2129--2170},
  year={2008},
  publisher={Wiley Online Library}
}

@article{marksaadDevelopmentTremorDetection2024,
  title = {Development of a {{Tremor Detection Algorithm}} for {{Use}} in an {{Academic Movement Disorders Center}}},
  author = {{Mark Saad} and Hefner, Sofia and Donovan, Suzann and Bernhard, Doug and Tripathi, Richa and Factor, Stewart A. and Powell, Jeanne M. and Kwon, Hyeokhyen and Sameni, Reza and Esper, Christine D. and McKay, J. Lucas},
  year = 2024,
  journal = {Sensors},
  volume = {24},
  number = {15},
  pages = {4960},
  issn = {1424-8220},
  doi = {10.3390/s24154960}
}

@article{ordonez_deep_2016,
author = {Ordóñez, Francisco Javier and Roggen, Daniel},
title = {{Deep Convolutional and LSTM Recurrent Neural Networks for Multimodal Wearable Activity Recognition}},
journal = {{Sensors}},
volume = {16},
year = {2016},
number = {1},
article-number = {115},
url = {https://www.mdpi.com/1424-8220/16/1/115},
PubMedID = {26797612},
ISSN = {1424-8220},
doi = {10.3390/s16010115},
pages = {115}
}

@article{vaswani2017attention,
  title={Attention is all you need},
  author={Vaswani, Ashish and Shazeer, Noam and Parmar, Niki and Uszkoreit, Jakob and Jones, Llion and Gomez, Aidan N and Kaiser, {\L}ukasz and Polosukhin, Illia},
  journal={Advances in neural information processing systems},
  volume={30},
  year={2017}
}

@article{friedrich2024video,
  title   = {Validation and application of computer vision algorithms for video-based tremor analysis},
  author  = {Friedrich, Maximilian U. and Roenn, Anna-Julia and Palmisano, Chiara and Alty, Jane and Paschen, Steffen and Deuschl, Guenther and Ip, Chi Wang and Volkmann, Jens and Muthuraman, Muthuraman and Peach, Robert and Reich, Martin M.},
  journal = {npj Digital Medicine},
  year    = {2024},
  volume  = {7},
  pages   = {165},
  doi     = {10.1038/s41746-024-01153-1},
  publisher = {Nature Publishing Group}
}

@article{haubenberger2016transducer,
  title   = {Transducer-Based Evaluation of Tremor},
  author  = {Haubenberger, Dietrich and Abbruzzese, Giovanni and Bain, Peter and Bajaj, Nin and Benito-Le{\'o}n, Juli{\'a}n and Bhatia, Kailash and Deuschl, G{\"u}nther and Forjaz, Mar{\'i}a J. and Hallett, Mark and Louis, Elan and Lyons, Kelly and Mestre, Tiago and Raethjen, Jan and Stamelou, Maria and Tan, Eng-King and Testa, Claudia and Elble, Rodger J.},
  journal = {Movement Disorders},
  year    = {2016},
  volume  = {31},
  number  = {9},
  pages   = {1327--1336},
  doi     = {10.1002/mds.26649},
  pmid    = {27273470},
  pmcid   = {PMC5014626}
}

@article{opitz2019macro,
  author        = {Juri Opitz and Sebastian Burst},
  title         = {Macro F1 and Macro F1},
  journal       = {CoRR},
  volume        = {abs/1911.03347},
  year          = {2021},
  doi           = {10.48550/arXiv.1911.03347},
  eprint        = {1911.03347},
  archivePrefix = {arXiv},
  primaryClass  = {cs.LG},
  url           = {https://arxiv.org/abs/1911.03347}
}

@article{gallagher2020savitzky,
  title={Savitzky-golay smoothing and differentiation filter},
  author={Gallagher, Neal B},
  journal={Eigenvector Research Incorporated},
  year={2020}
}

@inproceedings{carion2020end,
  title={End-to-end object detection with transformers},
  author={Carion, Nicolas and Massa, Francisco and Synnaeve, Gabriel and Usunier, Nicolas and Kirillov, Alexander and Zagoruyko, Sergey},
  booktitle={European conference on computer vision},
  pages={213--229},
  year={2020},
  organization={Springer}
}

@InProceedings{bach2015,
  title = 	 {Show, Attend and Tell: Neural Image Caption Generation with Visual Attention},
  author = 	 {Xu, Kelvin and Ba, Jimmy and Kiros, Ryan and Cho, Kyunghyun and Courville, Aaron and Salakhudinov, Ruslan and Zemel, Rich and Bengio, Yoshua},
  booktitle = 	 {Proceedings of the 32nd International Conference on Machine Learning},
  pages = 	 {2048--2057},
  year = 	 {2015},
  editor = 	 {Bach, Francis and Blei, David},
  volume = 	 {37},
  series = 	 {Proceedings of Machine Learning Research},
  address = 	 {Lille, France},
  month = 	 {07--09 Jul},
  publisher =    {PMLR},
  url = 	 {https://proceedings.mlr.press/v37/xuc15.html}
}

@article{mahendran2022differentiation,
  title={Differentiation of Parkinson’s disease tremor and essential tremor based on a novel hand posture},
  author={Mahendran, Sujitha and Bichsel, Oliver and Gassert, Roger and Baumann, Christian R and Imbach, Lukas L and Waldvogel, Daniel},
  journal={Clinical Parkinsonism \& Related Disorders},
  volume={7},
  pages={100146},
  year={2022},
  publisher={Elsevier}
}

@article{hssayeni2019,
AUTHOR = {Hssayeni, Murtadha D. and Jimenez-Shahed, Joohi and Burack, Michelle A. and Ghoraani, Behnaz},
TITLE = {Wearable Sensors for Estimation of Parkinsonian Tremor Severity during Free Body Movements},
JOURNAL = {Sensors},
VOLUME = {19},
YEAR = {2019},
NUMBER = {19},
ARTICLE-NUMBER = {4215},
URL = {https://www.mdpi.com/1424-8220/19/19/4215},
PubMedID = {31569335},
ISSN = {1424-8220},
DOI = {10.3390/s19194215}
}

@inproceedings{lundberg2017,
author = {Lundberg, Scott M. and Lee, Su-In},
title = {A unified approach to interpreting model predictions},
year = {2017},
isbn = {9781510860964},
publisher = {Curran Associates Inc.},
address = {Red Hook, NY, USA},
booktitle = {Proceedings of the 31st International Conference on Neural Information Processing Systems},
pages = {4768–4777},
numpages = {10},
location = {Long Beach, California, USA},
series = {NIPS'17}
}

@article{elble2016essential,
  title={The essential tremor rating assessment scale},
  author={Elble, Rodger J},
  journal={Journal of Neurology \& Neuromedicine},
  volume={1},
  number={4},
  year={2016}
}

@article{de_machine_2023,
  title = {Machine {{Learning}} in {{Tremor Analysis}}: {{Critique}} and {{Directions}}},
  shorttitle = {Machine {{Learning}} in {{Tremor Analysis}}},
  author = {De, Anwesan and Bhatia, Kailash P. and Volkmann, Jens and Peach, Robert and Schreglmann, Sebastian R.},
  year = {2023},
  journal = {Movement Disorders},
  volume = {38},
  number = {5},
  pages = {717--731},
  issn = {1531-8257},
  doi = {10.1002/mds.29376},
  urldate = {2025-04-27},
  langid = {english},
}

@article{kwon_explainable_2023,
  title = {An {{Explainable Spatial-Temporal Graphical Convolutional Network}} to {{Score Freezing}} of {{Gait}} in {{Parkinsonian Patients}}},
  author = {Kwon, Hyeokhyen and Clifford, Gari D. and Genias, Imari and Bernhard, Doug and Esper, Christine D. and Factor, Stewart A. and McKay, J. Lucas},
  year = {2023},
  month = {February},
  journal = {Sensors},
  volume = {23},
  number = {4},
  publisher = {MDPI},
  issn = {14248220},
  doi = {10.3390/s23041766},
  pmid = {36850363},
}

@article{bachlin_wearable_2010,
  title = {Wearable Assistant for {{Parkinsons}} Disease Patients with the Freezing of Gait Symptom},
  author = {B{\"a}chlin, Marc and Plotnik, Meir and Roggen, Daniel and Maidan, Inbal and Hausdorff, Jeffrey M. and Giladi, Nir and Tr{\"o}ster, Gerhard},
  year = {2010},
  month = {March},
  journal = {IEEE Transactions on Information Technology in Biomedicine},
  volume = {14},
  number = {2},
  pages = {436--446},
  issn = {10897771},
  doi = {10.1109/TITB.2009.2036165},
  urldate = {2024-06-16},
  pmid = {19906597},
}

@article{bulling_tutorial_2014,
  title = {A Tutorial on Human Activity Recognition Using Body-Worn Inertial Sensors},
  author = {Bulling, Andreas and Blanke, Ulf and Schiele, Bernt},
  year = {2014},
  month = {January},
  journal = {ACM Computing Surveys},
  volume = {46},
  number = {3},
  pages = {1--33},
  issn = {0360-0300, 1557-7341},
  doi = {10.1145/2499621},
  urldate = {2025-05-05},
  langid = {english}
}

@article{jeon2017high,
  title={High-accuracy automatic classification of Parkinsonian tremor severity using machine learning method},
  author={Jeon, Hyoseon and Lee, Woongwoo and Park, Hyeyoung and Lee, Hong Ji and Kim, Sang Kyong and Kim, Han Byul and Jeon, Beomseok and Park, Kwang Suk},
  journal={Physiological measurement},
  volume={38},
  number={11},
  pages={1980},
  year={2017},
  publisher={IOP Publishing}
}

@article{dosovitskiy2020image,
  title={An image is worth 16x16 words: Transformers for image recognition at scale},
  author={Dosovitskiy, Alexey},
  journal={arXiv preprint arXiv:2010.11929},
  year={2020}
}

@inproceedings{selvaraju2017grad,
  title={Grad-cam: Visual explanations from deep networks via gradient-based localization},
  author={Selvaraju, Ramprasaath R and Cogswell, Michael and Das, Abhishek and Vedantam, Ramakrishna and Parikh, Devi and Batra, Dhruv},
  booktitle={Proceedings of the IEEE international conference on computer vision},
  pages={618--626},
  year={2017}
}

@article{pham2017algorithm,
  title={Algorithm for turning detection and analysis validated under home-like conditions in patients with Parkinson’s disease and older adults using a 6 degree-of-freedom inertial measurement unit at the lower back},
  author={Pham, Minh H and Elshehabi, Morad and Haertner, Linda and Heger, Tanja and Hobert, Markus A and Faber, Gert S and Salkovic, Dina and Ferreira, Joaquim J and Berg, Daniela and Sanchez-Ferro, {\'A}lvaro and others},
  journal={Frontiers in neurology},
  volume={8},
  pages={135},
  year={2017},
  publisher={Frontiers Media SA}
}

@article{cappozzo1995position,
  title={Position and orientation in space of bones during movement: anatomical frame definition and determination},
  author={Cappozzo, Aurelio and Catani, Fabio and Della Croce, Ugo and Leardini, Alberto},
  journal={Clinical biomechanics},
  volume={10},
  number={4},
  pages={171--178},
  year={1995},
  publisher={Elsevier}
}

@article{samadi2020analysis,
  title={Analysis of hand tremor in parkinson’s disease: Frequency domain approach},
  author={Samadi, Elham and Ahmadi, Hessam and Rahatabad, Fereidoon Nowshiravan},
  journal={Frontiers in Biomedical Technologies},
  year={2020}
}

@article{thanawattano2015temporal,
  title={Temporal fluctuations of tremor signals from inertial sensor: a preliminary study in differentiating Parkinson’s disease from essential tremor},
  author={Thanawattano, Chusak and Pongthornseri, Ronachai and Anan, Chanawat and Dumnin, Songphon and Bhidayasiri, Roongroj},
  journal={Biomedical engineering online},
  volume={14},
  number={1},
  pages={101},
  year={2015},
  publisher={Springer}
}

@book{oppenheim1999discrete,
  title={Discrete-time signal processing},
  author={Oppenheim, Alan V},
  year={1999},
  publisher={Pearson Education India}
}

@article{beuter1999,
	author = {Beuter, Anne and Edwards, Roderick},
	journal = {Journal of Clinical Neurophysiology},
	doi = {10.1097/00004691-199909000-00010},
	issn = {0736-0258},
	number = {5},
	year = {1999},
	month = {9},
	pages = {484},
	publisher = {Ovid Technologies (Wolters Kluwer Health)},
	title = {Using {Frequency} {Domain} {Characteristics} to {Discriminate} {Physiologic} and {Parkinsonian} {Tremors}},
	url = {http://dx.doi.org/10.1097/00004691-199909000-00010},
	volume = {16},
}

@article{pang2020,
	author = {Pang, Yan and Christenson, Jake and Jiang, Feng and Lei, Tim and Rhoades, Remy and Kern, Drew and Thompson, John A. and Liu, Chao},
	journal = {Journal of Neuroscience Methods},
	doi = {10.1016/j.jneumeth.2019.108576},
	issn = {0165-0270},
	year = {2020},
	month = {3},
	pages = {108576},
	publisher = {Elsevier BV},
	title = {Automatic detection and quantification of hand movements toward development of an objective assessment of tremor and bradykinesia in {Parkinson}'s disease},
	url = {http://dx.doi.org/10.1016/j.jneumeth.2019.108576},
	volume = {333},
}

@article{luksys2018,
	author = {Luk{\v s}ys, Donatas and Jonaitis, Gintaras and Gri{\v s}kevi{\v c}ius, Julius},
	journal = {Parkinson's Disease},
	doi = {10.1155/2018/1683831},
	issn = {2090-8083},
	year = {2018},
	month = {jun 21},
	pages = {1--7},
	publisher = {Wiley},
	title = {Quantitative {Analysis} of {Parkinsonian} {Tremor} in a {Clinical} {Setting} {Using} {Inertial} {Measurement} {Units}},
	url = {http://dx.doi.org/10.1155/2018/1683831},
	volume = {2018},
}

@article{brittain2015,
	author = {Brittain, John-Stuart and Cagnan, Hayriye and Mehta, Arpan R. and Saifee, Tabish A. and Edwards, Mark J. and Brown, Peter},
	journal = {The Journal of Neuroscience},
	issn = {0270-6474},
	number = {2},
	year = {2015},
	month = {jan 14},
	pages = {795--806},
	publisher = {Society for Neuroscience},
	title = {Distinguishing the {Central} {Drive} to {Tremor} in {Parkinson}'s {Disease} and {Essential} {Tremor}},
	url = {http://dx.doi.org/10.1523/JNEUROSCI.3768-14.2015},
	volume = {35},
}

@article{meigal2012,
	author = {Meigal, A Yu and Rissanen, S M and Tarvainen, M P and Georgiadis, S D and Karjalainen, P A and Airaksinen, O and Kankaanp{\" a}{\" a}, M},
	journal = {Physiological Measurement},
	doi = {10.1088/0967-3334/33/3/395},
	issn = {0967-3334},
	number = {3},
	year = {2012},
	month = {feb 28},
	pages = {395--412},
	publisher = {IOP Publishing},
	title = {Linear and nonlinear tremor acceleration characteristics in patients with {Parkinson}'s disease},
	url = {http://dx.doi.org/10.1088/0967-3334/33/3/395},
	volume = {33},
}

@book{brown1997introduction,
  title={Introduction to Random Signals and Applied Kalman Filtering: With MATLAB Exercises and Solutions},
  author={Brown, Robert Grover and Hwang, Patrick Y.C.},
  year={1997},
  edition={3},
  publisher={Wiley},
}

@book{yates2025probability,
  title={Probability and Stochastic Processes: A Friendly Introduction for Electrical and Computer Engineers},
  author={Yates, Roy D. and Goodman, David J.},
  year={2025},
  edition={4},
  publisher={John Wiley \& Sons}
}

@incollection{suthaharan2016support,
  title={Support vector machine},
  author={Suthaharan, Shan},
  booktitle={Machine learning models and algorithms for big data classification: thinking with examples for effective learning},
  pages={207--235},
  year={2016},
  publisher={Springer}
}

@article{angelopoulos2023conformal,
  title={Conformal prediction: A gentle introduction},
  author={Angelopoulos, Anastasios N and Bates, Stephen},
  journal={Foundations and Trends in Machine Learning},
  volume={16},
  number={4},
  pages={494--591},
  year={2023},
  publisher={Emerald Publishing Limited}
}

@inproceedings{de2024impact,
  title={The Impact of Data Augmentation on Time Series Classification Models: An In-Depth Study with Biomedical Data},
  author={De, Bikram and Sakevych, Mykhailo and Metsis, Vangelis},
  booktitle={International Conference on Artificial Intelligence in Medicine},
  pages={192--203},
  year={2024},
  organization={Springer}
}

@article{huynh2022semi,
  title={Semi-supervised learning for medical image classification using imbalanced training data},
  author={Huynh, Tri and Nibali, Aiden and He, Zhen},
  journal={Computer methods and programs in biomedicine},
  volume={216},
  pages={106628},
  year={2022},
  publisher={Elsevier}
}

@article{gupta2020transfer,
  title={Transfer learning for clinical time series analysis using deep neural networks},
  author={Gupta, Priyanka and Malhotra, Pankaj and Narwariya, Jyoti and Vig, Lovekesh and Shroff, Gautam},
  journal={Journal of Healthcare Informatics Research},
  volume={4},
  number={2},
  pages={112--137},
  year={2020},
  publisher={Springer}
}

@article{liu2021efficient,
  title={Efficient training of visual transformers with small datasets},
  author={Liu, Yahui and Sangineto, Enver and Bi, Wei and Sebe, Nicu and Lepri, Bruno and Nadai, Marco},
  journal={Advances in Neural Information Processing Systems},
  volume={34},
  pages={23818--23830},
  year={2021}
}

@article{wen2022transformers,
  title={Transformers in time series: A survey},
  author={Wen, Qingsong and Zhou, Tian and Zhang, Chaoli and Chen, Weiqi and Ma, Ziqing and Yan, Junchi and Sun, Liang},
  journal={arXiv preprint arXiv:2202.07125},
  year={2022}
}

@article{shahtalebi2019wake,
  title={WAKE: Wavelet decomposition coupled with adaptive Kalman filtering for pathological tremor extraction},
  author={Shahtalebi, Soroosh and Atashzar, Seyed Farokh and Patel, Rajni V and Mohammadi, Arash},
  journal={Biomedical Signal Processing and Control},
  volume={48},
  pages={179--188},
  year={2019},
  publisher={Elsevier}
}

@inproceedings{jung2022imu,
  title={IMU Sensing Data-Based Kinetic Tremor Detection in Parkinson's Disease Patients},
  author={Jung, Woosub and Koltermann, Kenneth and Helm, Noah and Blackwell, GinaMari and Pretzer-Aboff, Ingrid and Cloud, Leslie and Zhou, Gang},
  booktitle={Proceedings of the 20th ACM Conference on Embedded Networked Sensor Systems},
  pages={772--773},
  year={2022}
}

@inproceedings{chen2025wavetremor,
  title={WaveTremor: Tremor Detection for Parkinson's Disease via Spatial-Temporal Learning},
  author={Chen, Xinyu and Koltermann, Kenneth and Clapham, John and Blackwell, Ginamari and Cloud, Leslie and Pretzer-Aboff, Ingrid and Shao, Huajie and Zhou, Gang},
  booktitle={Proceedings of the ACM/IEEE International Conference on Connected Health: Applications, Systems and Engineering Technologies},
  pages={309--313},
  year={2025}
}

@article{raethjen2004tremor,
  title={Tremor analysis in two normal cohorts},
  author={Raethjen, J and Lauk, M and K{\"o}ster, B and Fietzek, U and Friege, L and Timmer, J and L{\"u}cking, CH and Deuschl, G},
  journal={Clinical Neurophysiology},
  volume={115},
  number={9},
  pages={2151--2156},
  year={2004},
  publisher={Elsevier}
}

@article{deuschl1995tremor,
  title={Tremor classification and tremor time series analysis},
  author={Deuschl, G{\"u}nther and Lauk, Michael and Timmer, Jens},
  journal={Chaos: An Interdisciplinary Journal of Nonlinear Science},
  volume={5},
  number={1},
  pages={48--51},
  year={1995},
  publisher={American Institute of Physics}
}

@article{timmer2000cross,
  title={Cross-spectral analysis of tremor time series},
  author={Timmer, Jens and Lauk, Michael and H{\"a}u{\ss}ler, S and Radt, V and K{\"o}ster, B and Hellwig, B and Guschlbauer, B and L{\"u}cking, CH and Eichler, M and Deuschl, G{\"u}nther},
  journal={international Journal of Bifurcation and chaos},
  volume={10},
  number={11},
  pages={2595--2610},
  year={2000},
  publisher={World Scientific}
}

@article{ribeiro2019bag,
  title={Bag of samplings for computer-assisted Parkinson's disease diagnosis based on recurrent neural networks},
  author={Ribeiro, Luiz CF and Afonso, Luis CS and Papa, Joao P},
  journal={Computers in biology and medicine},
  volume={115},
  pages={103477},
  year={2019},
  publisher={Elsevier}
}

@article{kim2018wrist,
  title={Wrist sensor-based tremor severity quantification in Parkinson's disease using convolutional neural network},
  author={Kim, Han Byul and Lee, Woong Woo and Kim, Aryun and Lee, Hong Ji and Park, Hye Young and Jeon, Hyo Seon and Kim, Sang Kyong and Jeon, Beomseok and Park, Kwang S},
  journal={Computers in biology and medicine},
  volume={95},
  pages={140--146},
  year={2018},
  publisher={Elsevier}
}

@article{sigcha2021automatic,
  title={Automatic resting tremor assessment in Parkinson’s disease using smartwatches and multitask convolutional neural networks},
  author={Sigcha, Luis and Pav{\'o}n, Ignacio and Costa, N{\'e}lson and Costa, Susana and Gago, Miguel and Arezes, Pedro and L{\'o}pez, Juan Manuel and De Arcas, Guillermo},
  journal={Sensors},
  volume={21},
  number={1},
  pages={291},
  year={2021},
  publisher={MDPI}
}

@article{san2020parkinson,
  title={Parkinson’s disease tremor detection in the wild using wearable accelerometers},
  author={San-Segundo, Rub{\'e}n and Zhang, Ada and Cebulla, Alexander and Panev, Stanislav and Tabor, Griffin and Stebbins, Katelyn and Massa, Robyn E and Whitford, Andrew and De la Torre, Fernando and Hodgins, Jessica},
  journal={Sensors},
  volume={20},
  number={20},
  pages={5817},
  year={2020},
  publisher={MDPI}
}

@article{shahtalebi2020phtnet,
  title={PHTNet: Characterization and deep mining of involuntary pathological hand tremor using recurrent neural network models},
  author={Shahtalebi, Soroosh and Atashzar, Seyed Farokh and Samotus, Olivia and Patel, Rajni V and Jog, Mandar S and Mohammadi, Arash},
  journal={Scientific reports},
  volume={10},
  number={1},
  pages={2195},
  year={2020},
  publisher={Nature Publishing Group UK London}
}

@article{rajkomar2018scalable,
  title={Scalable and accurate deep learning with electronic health records},
  author={Rajkomar, Alvin and Oren, Eyal and Chen, Kai and Dai, Andrew M and Hajaj, Nissan and Hardt, Michaela and Liu, Peter J and Liu, Xiaobing and Marcus, Jake and Sun, Mimi and others},
  journal={NPJ digital medicine},
  volume={1},
  number={1},
  pages={18},
  year={2018},
  publisher={Nature Publishing Group UK London}
}

@article{esteva2019guide,
  title={A guide to deep learning in healthcare},
  author={Esteva, Andre and Robicquet, Alexandre and Ramsundar, Bharath and Kuleshov, Volodymyr and DePristo, Mark and Chou, Katherine and Cui, Claire and Corrado, Greg and Thrun, Sebastian and Dean, Jeff},
  journal={Nature medicine},
  volume={25},
  number={1},
  pages={24--29},
  year={2019},
  publisher={Nature Publishing Group US New York}
}

@article{shahtalebi2021deep,
  title={A deep explainable artificial intelligent framework for neurological disorders discrimination},
  author={Shahtalebi, Soroosh and Atashzar, S Farokh and Patel, Rajni V and Jog, Mandar S and Mohammadi, Arash},
  journal={Scientific reports},
  volume={11},
  number={1},
  pages={9630},
  year={2021},
  publisher={Nature Publishing Group UK London}
}

@article{welch1967use,
  title={The use of fast Fourier transform for the estimation of power spectra: A method based on time averaging over short, modified periodograms},
  author={Welch, Peter},
  journal={IEEE Transactions on audio and electroacoustics},
  volume={15},
  number={2},
  pages={70--73},
  year={1967},
  publisher={IEEE}
}

@article{pulliam2017continuous,
  title={Continuous assessment of levodopa response in Parkinson's disease using wearable motion sensors},
  author={Pulliam, Christopher L and Heldman, Dustin A and Brokaw, Elizabeth B and Mera, Thomas O and Mari, Zoltan K and Burack, Michelle A},
  journal={IEEE Transactions on Biomedical Engineering},
  volume={65},
  number={1},
  pages={159--164},
  year={2017},
  publisher={IEEE}
}

@article{heldman2014clinician,
  title={Clinician versus machine: reliability and responsiveness of motor endpoints in Parkinson's disease},
  author={Heldman, Dustin A and Espay, Alberto J and LeWitt, Peter A and Giuffrida, Joseph P},
  journal={Parkinsonism \& related disorders},
  volume={20},
  number={6},
  pages={590--595},
  year={2014},
  publisher={Elsevier}
}

@article{dorsey2018parkinson,
  title={The Parkinson pandemic—a call to action},
  author={Dorsey, E Ray and Bloem, Bastiaan R},
  journal={JAMA neurology},
  volume={75},
  number={1},
  pages={9--10},
  year={2018}
}

@article{maetzler2016clinical,
  title={A clinical view on the development of technology-based tools in managing Parkinson's disease},
  author={Maetzler, Walter and Klucken, Jochen and Horne, Malcolm},
  journal={Movement Disorders},
  volume={31},
  number={9},
  pages={1263--1271},
  year={2016},
  publisher={Wiley Online Library}
}

@incollection{wilcoxon1992individual,
  title={Individual comparisons by ranking methods},
  author={Wilcoxon, Frank},
  booktitle={Breakthroughs in statistics: Methodology and distribution},
  pages={196--202},
  year={1992},
  publisher={Springer}
}

@article{wainer2021nested,
  title={Nested cross-validation when selecting classifiers is overzealous for most practical applications},
  author={Wainer, Jacques and Cawley, Gavin},
  journal={Expert Systems with Applications},
  volume={182},
  pages={115222},
  year={2021},
  publisher={Elsevier}
}

@article{parvandeh2020consensus,
  title={Consensus features nested cross-validation},
  author={Parvandeh, Saeid and Yeh, Hung-Wen and Paulus, Martin P and McKinney, Brett A},
  journal={Bioinformatics},
  volume={36},
  number={10},
  pages={3093--3098},
  year={2020},
  publisher={Oxford University Press}
}

\section*{Acknowledgments}

\section*{Supporting information}
\renewcommand{\thetable}{S\arabic{table}}
\setcounter{table}{0}
\renewcommand{\thefigure}{S\arabic{figure}}
\setcounter{figure}{0}
\begin{figure*}[h!]
\centering
\includegraphics[width=\textwidth]{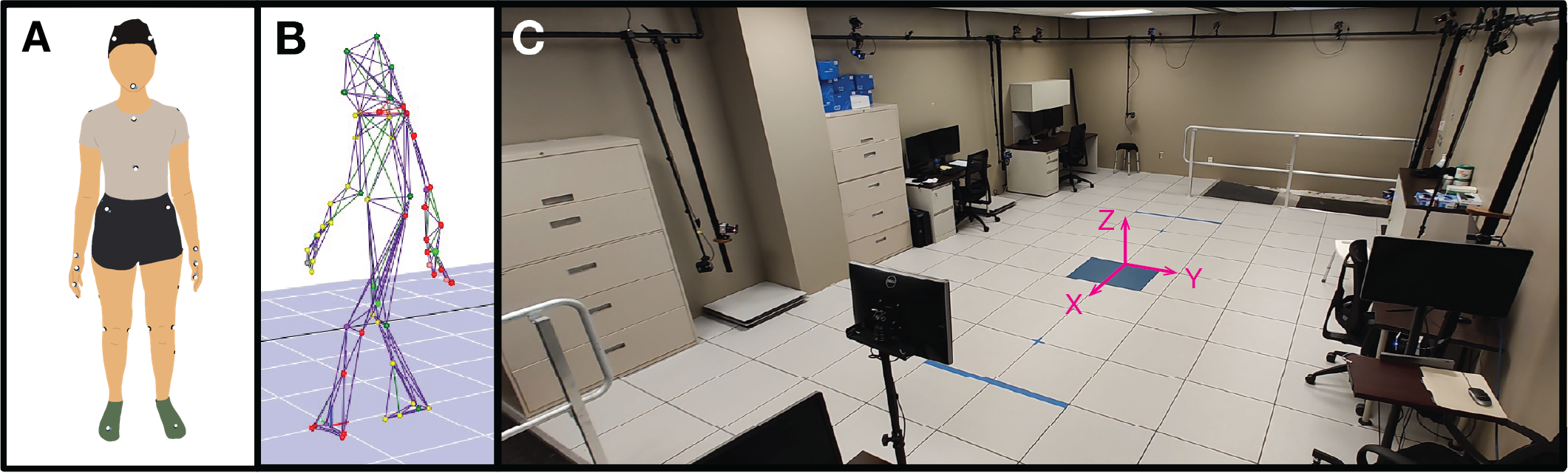}
\caption{
This figure illustrates marker placement and motion capture facility as described in Saad \textit{et al. }~\cite{marksaadDevelopmentTremorDetection2024}.
(\textbf{A}) Figure illustrating the marker placement on the 16 body extremities of a study participant. 
(\textbf{B}) De-identified ``wire frame" representation of the participant in \textbf{A}.
(\textbf{C})The origin of the kinematic coordinate system is superimposed in the clinical motion capture facility.
}
\label{fig:marker_placement}
\end{figure*}
\begin{table}[htbp]
\centering
\caption{Kinematic marker descriptions for markers on the trunk. Markers that appear on both sides of the body are listed for the right side only and are coded beginning with "R". Replacing this character with "L" will designate the corresponding marker on the left side of the body.}
\label{tab:trunk_markers}
\begin{tabular}{lll}
\hline
\textbf{Extremity} & \textbf{Marker Code} & \textbf{Description} \\
\hline
Head & Front\_Head & Center of forehead, on cap \\
 & JAW & Mental protuberance \\
 & RBHD & Right back head, on cap \\
 & RFHD & Right front head, on cap \\
 & Rear\_Head & Rear of head, on cap \\
 & TopHead & Top of head \\
 & Top\_Head & Top of head \\
\hline
Shoulders & C7 & Seventh cervical vertebra \\
 & RBAK & Right scapula (asymmetry marker) \\
 & RSHO & Right acromioclavicular joint \\
 & R\_Shoulder & Right acromion process \\
 & STRN & Xiphoid process \\
\hline
Pelvis & LASI & Left anterior superior iliac spine \\
 & RASI & Right anterior superior iliac spine \\
 & RIC & Right iliac crest \\
 & RPSI & Right posterior superior iliac spine \\
 & R\_ASIS & Right anterior superior iliac spine \\
 & V\_Sacral & Sacrum \\
\hline
Thorax & CLAV & Clavicular notch \\
 & R\_Clavicle & Right clavicle \\
 & R\_Scap\_Inf & Right scapula inferior angle \\
 & R\_Scapula & Right supraspinous fossa \\
 & T10 & 10th thoracic vertebra \\
\hline
\end{tabular}
\end{table}
\begin{table}[htbp]
\centering
\caption{Kinematic marker descriptions for markers on the arms. Markers that appear on both sides of the body are listed for the right side only and are coded beginning with "R". Replacing this character with "L" will designate the corresponding marker on the left side of the body.}
\label{tab:arms_markers}
\begin{tabular}{lll}
\hline
\textbf{Extremity} & \textbf{Marker Code} & \textbf{Description} \\
\hline
R\_Dist\_Arm & RFRM & Lateral surface of forearm \\
 & RWRA & Radial side of wrist \\
 & RWRB & Ulnar side of wrist \\
 & R\_Forearm & Lateral surface of forearm \\
 & R\_Radius & Right styloid process of radius \\
 & R\_Ulna & Mid region of ulna \\
\hline
R\_Hand & RFIN & Third finger, first metacarpal joint \\
 & RFINGM2 & Third finger, second metacarpal joint \\
 & RFINGM3 & Third finger, most distal segment \\
 & RTHM1 & Thumb, first metacarpal \\
 & RTHM2 & Thumb, second metacarpal \\
 & RTHM3 & Thumb, most distal segment \\
 & R\_Finger3\_M1 & Third finger, first metacarpal joint \\
 & R\_Finger3\_M2 & Third finger, second metacarpal joint \\
 & R\_Finger3\_M3 & Third finger, most distal segment \\
 & R\_Hand & Radial surface of wrist \\
 & R\_Thumb\_M1 & Thumb, first metacarpal \\
 & R\_Thumb\_M2 & Thumb, second metacarpal \\
 & R\_Thumb\_M3 & Thumb, most distal segment \\
 & R\_Wrist & Radial surface of wrist \\
\hline
R\_Prox\_Arm & RELB & Right lateral epicondyle \\
 & R\_BicepsLateral & Lateral surface of upper arm \\
 & R\_Biceps\_Lateral & Lateral surface of upper arm \\
 & R\_Elbow & Right lateral epicondyle \\
 & R\_Elbow\_Medial & Right medial epicondyle \\
\hline
\end{tabular}
\end{table}
\begin{table}[htbp]
\centering
\caption{Kinematic marker descriptions for markers on the legs. Markers that appear on both sides of the body are listed for the right side only and are coded beginning with "R". 
Replacing this character with "L" will designate the corresponding marker on the left side of the body.}
\label{tab:legs_markers}
\begin{tabular}{lll}
\hline
\textbf{Extremity} & \textbf{Marker Code} & \textbf{Description} \\
\hline
R\_Dist\_Leg & RANK & Lateral aspect of ankle \\
 & RANKM & Medial aspect of ankle \\
 & RTIB & Midpoint of tibia \\
 & R\_Ankle & Lateral aspect of ankle \\
 & R\_Ankle\_Medial & Medial aspect of ankle \\
 & R\_Shank & Midpoint of tibia \\
\hline
R\_Foot & RFTM & Dorsal/medial surface of foot midway \\
 &  & between ankle and toe \\
 & RHEE & Distal surface of heel \\
 & RTOE & Third metatarsal \\
 & R\_Hallux & Dorsal surface of big toe \\
 & R\_Heel & Distal surface of heel \\
 & R\_MedFoot & Dorsal/medial surface of foot midway \\
 &  & between ankle and toe \\
\hline
R\_Prox\_Leg & R\_Toe & Third metatarsal \\
 & RKNE & Lateral aspect of flexion--extension axis \\
 &  & of knee \\
 & RKNEM & Medial aspect of flexion--extension axis \\
 &  & of knee \\
 & RTHI & Upper lateral 1/3 surface of thigh \\
 & R\_Knee & Lateral aspect of flexion--extension axis \\
 &  & of knee \\
 & R\_Knee\_Medial & Medial aspect of flexion--extension axis \\
 &  & of knee \\
\hline
\end{tabular}
\end{table}
\begin{table*}[!ht]
    \centering
    \caption{Frequency distribution of tremor labels for each body part extremity in the Saad dataset~\cite{marksaadDevelopmentTremorDetection2024}.
    L refers to Left, R refers to Right. 
    } 
    \begin{adjustbox}{max width=\linewidth}
    \begin{tabular}{lcc}
        \hline
        \\
        Body Part & Absent & Present
        \\ \\
        \hline
        Head & 76 & 52 \\
        Shoulders & 71 & 1 \\
        Thorax & 78 & 1 \\
        Pelvis & 71 & 1 \\
        L\_Dist\_Arm & 73 & 5 \\
        R\_Dist\_Arm & 72 & 1 \\
        L\_Hand & 237 & 280 \\
        R\_Hand & 242 & 279 \\
        L\_Prox\_Arm & 71 & 3 \\
        R\_Prox\_Arm & 72 & 1 \\
        L\_Dist\_Leg & 81 & 21 \\
        R\_Dist\_Leg & 8 & 8 \\
        L\_Foot & 76 & 30 \\
        R\_Foot & 81 & 15 \\
        L\_Prox\_Leg & 81 & 21 \\
        R\_Prox\_Leg & 81 & 8 \\
        \hline
    \end{tabular}    
    \end{adjustbox}
    \label{tab:label_distribution}
\end{table*}
\begin{table*}[!ht]
    \centering
    \caption{The number of kinematic markers in each extremity in the Saad dataset~\cite{marksaadDevelopmentTremorDetection2024}.
      L refers to Left, R refers to Right.
    }
    \begin{adjustbox}{max width=\linewidth}
    \begin{tabular}{lc}
        \hline
        \\
        Body Part & Number of Markers
        \\ \\
        \hline
        Head & 12 \\
        Shoulders & 9 \\
        Thorax & 15 \\
        Pelvis & 9 \\
        L\_Dist\_Arm & 9 \\
        R\_Dist\_Arm & 9 \\
        L\_Hand & 21 \\
        R\_Hand & 21 \\
        L\_Prox\_Arm & 9 \\
        R\_Prox\_Arm & 9 \\
        L\_Dist\_Leg & 9 \\
        R\_Dist\_Leg & 9 \\
        L\_Foot & 9 \\
        R\_Foot & 9 \\
        L\_Prox\_Leg & 9 \\
        R\_Prox\_Leg & 9 \\
        \hline
    \end{tabular}    
    \end{adjustbox}
    \label{tab:num_kinematic_markers}
\end{table*}

\begin{figure*}[!ht]
\centering
\begin{overpic}[width=\linewidth]{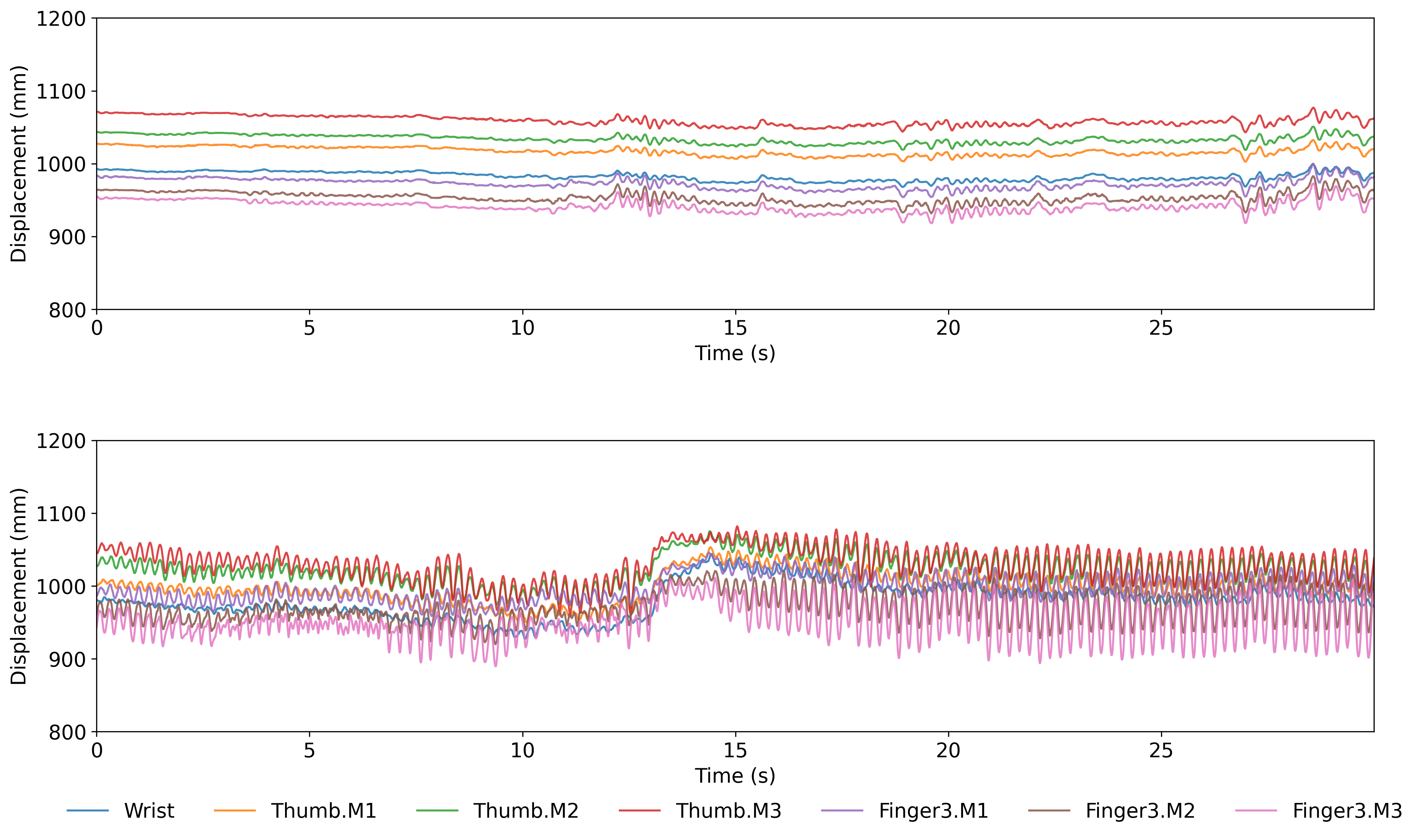}
    \put(-5,60){\colorbox{white}{\large\textsf{\textbf{A}}}}
    \put(-5,30){\colorbox{white}{\large\textsf{\textbf{B}}}}
\end{overpic}

\caption{
Intermittent versus sustained tremor during the sit-UEopp task (seated with arms in a T-pose parallel to the ground, fingers opposed).
The raw clinician annotations were \textit{``littler intermittent''} for the intermittent example (A) and \textit{``significant''} for the sustained example (B). 
In (A), tremor patterns are observed during short time periods during the trial, i.e., 12--14\,s, 18--22\,s, and 26--28\,s. 
In (B), tremor patterns are observed throughout the session.}
\label{fig:intermittent_sustained}
\end{figure*}

\end{document}